\documentclass[3p,12pt]{elsarticle}
\usepackage{lipsum}
\makeatletter
\def\ps@pprintTitle{%
	\let\@oddhead\@empty
	\let\@evenhead\@empty
	\def\@oddfoot{}%
	\let\@evenfoot\@oddfoot}
\makeatother

\usepackage{amssymb}
\usepackage{amsmath}

\usepackage{xcolor}

\journal{Extreme Mechanics Letters}

\begin{document}
	
\begin{frontmatter}	
	\title{Modeling the Dynamics of Sub-Millisecond Electroadhesive Engagement and Release Times}
	
	\author{Ahad M. Rauf}
	\ead{ahadrauf@stanford.edu}
	\author{Sean Follmer}
	\affiliation{organization={Department of Mechanical Engineering at Stanford University}, city={Stanford}, state={CA 94305}, country={USA}}

	\begin{abstract}
		Electroadhesive clutches are electrically controllable switchable adhesives commonly used in soft robots and haptic user interfaces. They can form strong bonds to a wide variety of surfaces at low power consumption. However, electroadhesive clutches in the literature engage to and release from substrates several orders of magnitude slower than a traditional electrostatic model would predict. Large release times, in particular, can limit electroadhesion's usefulness in high-bandwidth applications. We develop a novel electromechanical model for electroadhesion, factoring in polarization dynamics, the drive circuitry's rise and fall times, and contact mechanics between the dielectric and substrate. We show in simulation and experimentally how different design parameters affect the engagement and release times of centimeter-scale electroadhesive clutches to metallic substrates, and we find that the model accurately captures the magnitude and trends of our experimental results. In particular, we find that higher drive frequencies, narrower substrate aspect ratios, and faster drive circuitry output stages enable significantly faster release times. The fastest clutches have engagement times less than 15 \textmu s and release times less than 875 \textmu s, which are 10$\times$ and 17.1$\times$ faster, respectively, than the best times found in prior literature on centimeter-scale electroadhesive clutches.
	\end{abstract}
	
%
	
	
\end{frontmatter}

\section{Introduction} \label{sec:introduction}

Switchable adhesives can dynamically adhere to and release from contacting surfaces. They are widely used in soft robots, haptic interfaces, and biomedical devices, with applications in robotic grippers \cite{Cacucciolo_Shea_Carbone_2022, Mastrangelo_Caruso_Carbone_Cacucciolo_2023}, wall climbing and perching robots \cite{Park_Drew_Follmer_Rivas-Davila_2020, Graule_Chirarattananon_Fuller_Jafferis_Ma_Spenko_Kornbluh_Wood_2016, Guo_Guo_Liu_Liu_Leng_2022}, exoskeletons \cite{Krimsky_Collins_2024, Diller_Collins_Majidi_2018}, haptic user interfaces \cite{Rauf_Bernardo_Follmer_2023, Zhang_Gonzalez_Guo_Follmer_2019, Leithinger_Zhou_Acome_Rauf_Han_Shultz_Mullenbach_2023, Shultz_Peshkin_Colgate_2018}, and kinesthetic garments \cite{Vechev_Hinchet_Coros_Thomaszewski_Hilliges_2022, Hinchet_Shea_2022}. Different switchable adhesion mechanisms can be compared by their shear force capacity when gripping different substrates and by their engagement and release speeds \cite{Croll_Hosseini_Bartlett_2019}. Mechanisms with high shear force capacities, like granular jamming actuators and shape memory polymers, often require long switching times due to viscoelastic effects, while faster adhesives like liquid crystal elastomers and gecko-inspired adhesives often have caveats like low force capacity or poor adhesion to rough surfaces \cite{Wang_Chortos_2022, Han_Hajj-Ahmad_Cutkosky_2020}.

\begin{figure*}[!t]
	\centering
	\includegraphics[width=\linewidth]{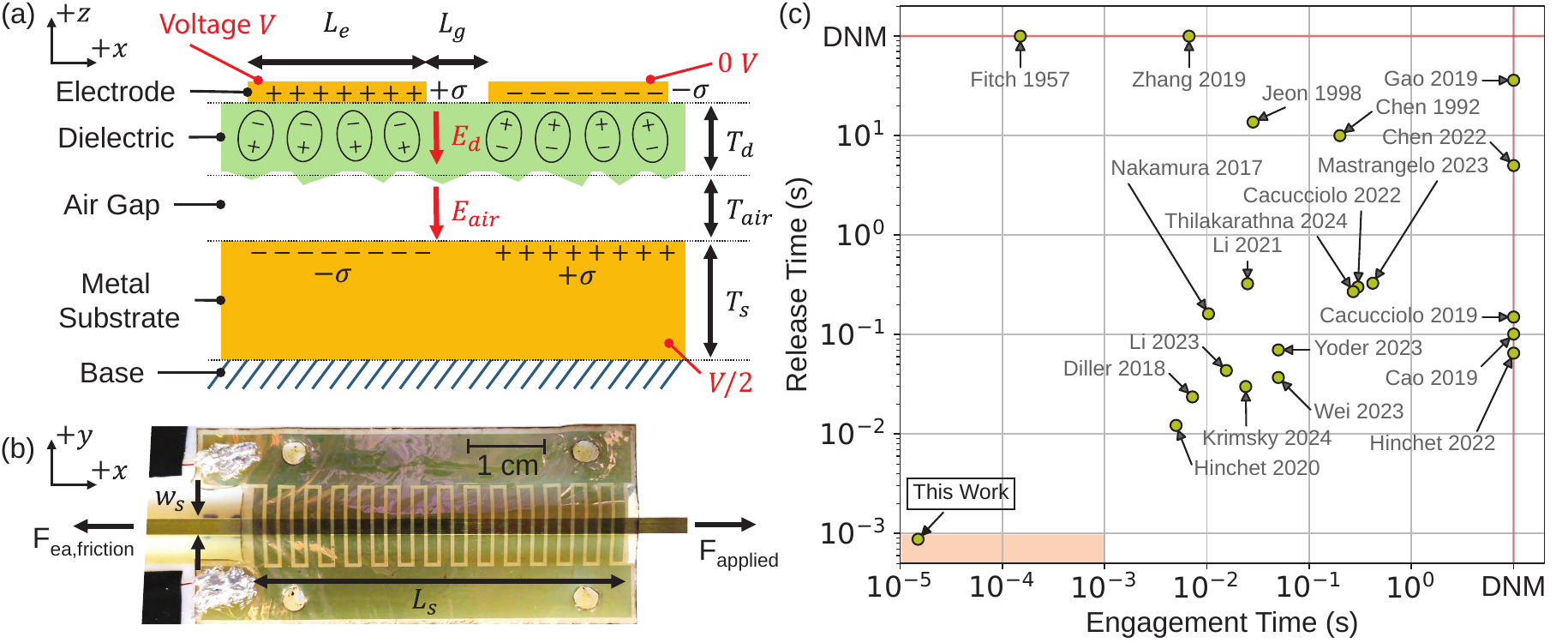}
	\caption{(a) Schematic diagram and (b) picture of an electroadhesive clutch. When a voltage $V$ is applied, the polarization of bound charges in the dielectric amplifies the field strength, attracting the film to the metal substrate. The resulting friction can counteract applied shear loads on the substrate. (c) Comparison of engagement and release times of centimeter-scale EA clutches in the literature. Papers that did not report one of the two metrics are placed on the red ``Did Not Measure'' (DNM) axis. The orange highlighted area indicates sub-millisecond engagement and release times. The raw data is provided in Supplementary Materials, Table S1. \cite{Cacucciolo_Shea_Carbone_2022, Mastrangelo_Caruso_Carbone_Cacucciolo_2023, Krimsky_Collins_2024, Diller_Collins_Majidi_2018, Zhang_Gonzalez_Guo_Follmer_2019, Hinchet_Shea_2022, Chen_Liu_Wang_Zhu_Tao_Zhang_Jiang_2022, Nakamura_Yamamoto_2017, Hinchet_Shea_2020, Fitch_1957, Jeon_Higuchi_1998, Gao_Cao_Guo_Conn_2019, Chen_Sarhadi_1992, Thilakarathna_Phlernjai_2024, Li_2021, Cacucciolo_Shintake_Shea_2019, Yoder_Macari_Kleinwaks_Schmidt_Acome_Keplinger_2023, Cao_Gao_Guo_Conn_2019, Wei_Xiong_Dong_Wang_Liang_Tang_Xu_Wang_Wang_2023, Li_Xiong_Ma_Yang_Ma_Tao_2023}}
	\label{fig:schematic_diagram}
\end{figure*}

Electroadhesion (EA), or the electrostatic attraction between two contacting materials separated by a dielectric and held at a potential difference, has emerged as an effective electrically triggered switchable adhesive. Its scalable manufacturing process, high shear force capacity against a wide range of substrates, relatively fast actuation speeds, low power consumption, and low profile have made EA useful for space- and power-constrained applications. In this paper, we focus on a common implementation, shown in Fig. \ref{fig:schematic_diagram}(a), where one of the contacting materials is a dielectric patterned with interdigitated comb electrodes while the other contacting material can be any conductive or insulating substrate. This form factor, originally developed in 1968 by Krape \cite{Krape_1968}, sacrifices some adhesive force capacity compared to a parallel plate configuration in which both contacting materials have electrodes. However, EA with Interdigitated Co-planar Electrodes (shortened to EA-ICE in this paper) allow all the electronics to be concentrated into a single surface, reducing bulk and enabling the clutch to adhere to small, delicate objects and generic surfaces in the field.

Despite electroadhesion's widespread utility, there is a lack of accurate dynamics models for how quickly EA clutches can adhere to and release from surfaces. Prior work has shown that the dielectrics themselves can be charged and discharged on the order of microseconds \cite{Wang_Lu_Lanagan_Zhang_2009, Tang_Lin_Sodano_2012, AliAbbasi_Martinsen_Pettersen_Colgate_Basdogan_2024}, and sub-millimeter-scale electroadhesive systems can indeed measure engagement and release times on this order of magnitude \cite{Hays_1991, Rauf_Contreras_Shih_Schindler_Pister_2022, Contreras_Pister_2017, Shih_Contreras_Massey_Greenspun_Pister_2018}. However, as shown in Fig. \ref{fig:schematic_diagram}(c), every paper we found on centimeter-scale electroadhesive clutches reported engagement times from hundreds of microseconds to milliseconds and release times ranging from tens of milliseconds to minutes. While some models predicting minute-scale release times have been proposed (although, to our knowledge, never compared numerically against experimental data) \cite{Chen_Liu_Wang_Zhu_Tao_Zhang_Jiang_2022, Nakamura_Yamamoto_2017, Chen_Zhang_Song_Fang_Sindersberger_Monkman_Guo_2020, Rajagopalan_Muthu_Liu_Luo_Wang_Wan_2022}, experimental results in the microsecond to millisecond range have largely gone unexplained. Slow release times limit EA's scalability for high-bandwidth applications \cite{Shultz_Peshkin_Colgate_2018, Leroy_Shea_2023}. Slow dynamics also leads to ``stick-slip'' behavior, where after an applied load exceeds an EA clutch's force capacity the clutch slips, losing up to 70\% of its frictional capacity and requiring tens of milliseconds to re-engage \cite{Diller_Collins_Majidi_2018, Hinchet_Shea_2020, Chen_Bergbreiter_2017}. In this paper, we aim to understand and model this literature gap, opening optimization potential for EA pads to serve as ultra-fast, electrically controllable, planar adhesive clutches.

We developed an electromechanical EA dynamics model, factoring in the dynamics of the drive circuitry, polarization, and mechanical contact between the EA pad and substrate. These considerations enable our model, unlike prior EA dynamics models, to predict microsecond- to millisecond-scale engagement times and millisecond-scale release times in line with the range seen in previous experimental studies. We test using simulation and experiments how different geometric and operating parameters affect the EA clutch's engagement and release times to metallic substrates when loaded in shear. The fastest clutches are demonstrated to have engagement times less than 15 \textmu s and release times less than 875 \textmu s, which are 10$\times$ and 17.1$\times$ faster, respectively, than the best times found in prior literature on centimeter-scale EA clutches. These improvements, particularly to release time, enable our EA clutches' frequency responses to have -3 dB bandwidths over 10 kHz, and we also show that EA clutches tested in conditions that result in better release times also often recover faster after a slip. We anticipate these results can be used to optimize EA clutches for high-bandwidth switchable adhesion applications, such as haptic interfaces, wearable robots, and biomedical devices. 

\section{Modeling Electroadhesion} \label{sec:modeling_electroadhesion}
We first modeled the electrical and mechanical dynamics separately to understand the relevant design variables, and then combined them into a full electromechanical dynamics model. While our derivations focus on EA-ICE clutches adhering to metallic substrates, the final equations are almost identical for parallel plate EA clutches, as shown in Supplementary Materials, Sec. S2.

\subsection{Quasistatic Electroadhesive Force} \label{subsec:electrostatic_force}
As shown in Fig. \ref{fig:schematic_diagram}(a, b), we consider an interdigitated comb pattern of $N$ electrodes, each with length $L_e$ and spaced $L_g$ apart. The EA pad is attracted to a substrate with overlap length $L_s$, width $w_s$, and thickness $T_s$. The dielectric thickness is $T_d$, its relative permittivity is $\kappa$, and the average air gap is $T_{air}$. We assume a metal substrate to facilitate modeling, as discussed further in Sec. \ref{subsec:electrostatic_dynamics}. Fig. \ref{fig:schematic_diagram}(a) shows a diagram of the layer stack's electric fields and charge densities. After depositing a surface charge density $\sigma$ on an electrode with overlap area $A=L_ew_s$, a charge $-\sigma$ is induced in the substrate. Inspired by Nakamura and Yamamoto \cite{Nakamura_Yamamoto_2017}, the capacitance between each electrode and the substrate can be modeled as a dielectric and air capacitor in series. Assuming for now that both the electrode and substrate surfaces are perfectly flat, the capacitance between $N$ adjacent electrodes (half held at potential $V$, half at ground) is thus:
\begin{equation} \label{eq:capacitance}
	C = \frac{N\kappa\varepsilon_0 A}{4(T_d + \kappa T_{air})}
\end{equation}

Rearranging this equation, we can estimate the air gap via:
\begin{equation} \label{eq:t_air_from_capacitance}
	T_{air} = \frac{N\varepsilon_0 A}{4C} - \frac{T_d}{\kappa}
\end{equation}

The electric fields in the dielectric and air are $E_d = \frac{\sigma}{\kappa \varepsilon_0}$ and $E_{air} = \frac{\sigma}{\varepsilon_0}$, respectively. Because the metal substrate is left floating, its effective potential is $V/2$. This potential difference can be related to the electric fields and charge densities by:
\begin{equation}
	\frac{V}{2} =  E_dT_d + E_{air}T_{air} = \sigma \left( \frac{T_d}{\kappa \varepsilon_0} + \frac{T_{air}}{\varepsilon_0} \right)
\end{equation}

The normal force on the substrate from an electroadhesive pad with $N$ electrodes is thus:
\begin{equation} \label{eq:electroadhesive_normal_force}
	F_{ea}(T_{air}) = \frac{\sigma NA E_{air}}{2} = \frac{\kappa^2\varepsilon_0}{2} \frac{NA(V/2)^2}{(T_d + \kappa T_{air})^2}
\end{equation}

Eq. \ref{eq:electroadhesive_normal_force} differs from a conventional parallel plate force derivation that ignores the effect of the air gap \cite{Rauf_Bernardo_Follmer_2023, Chen_Bergbreiter_2017}:
\begin{equation} \label{eq:force_electrostatic_noairgap}
	F_{es,T_{air} = 0} = \frac{\kappa\varepsilon_0}{2} \frac{NA(V/2)^2}{T_d^2}
\end{equation}

\begin{figure*}[t]
	\centering
	\includegraphics[width=\linewidth]{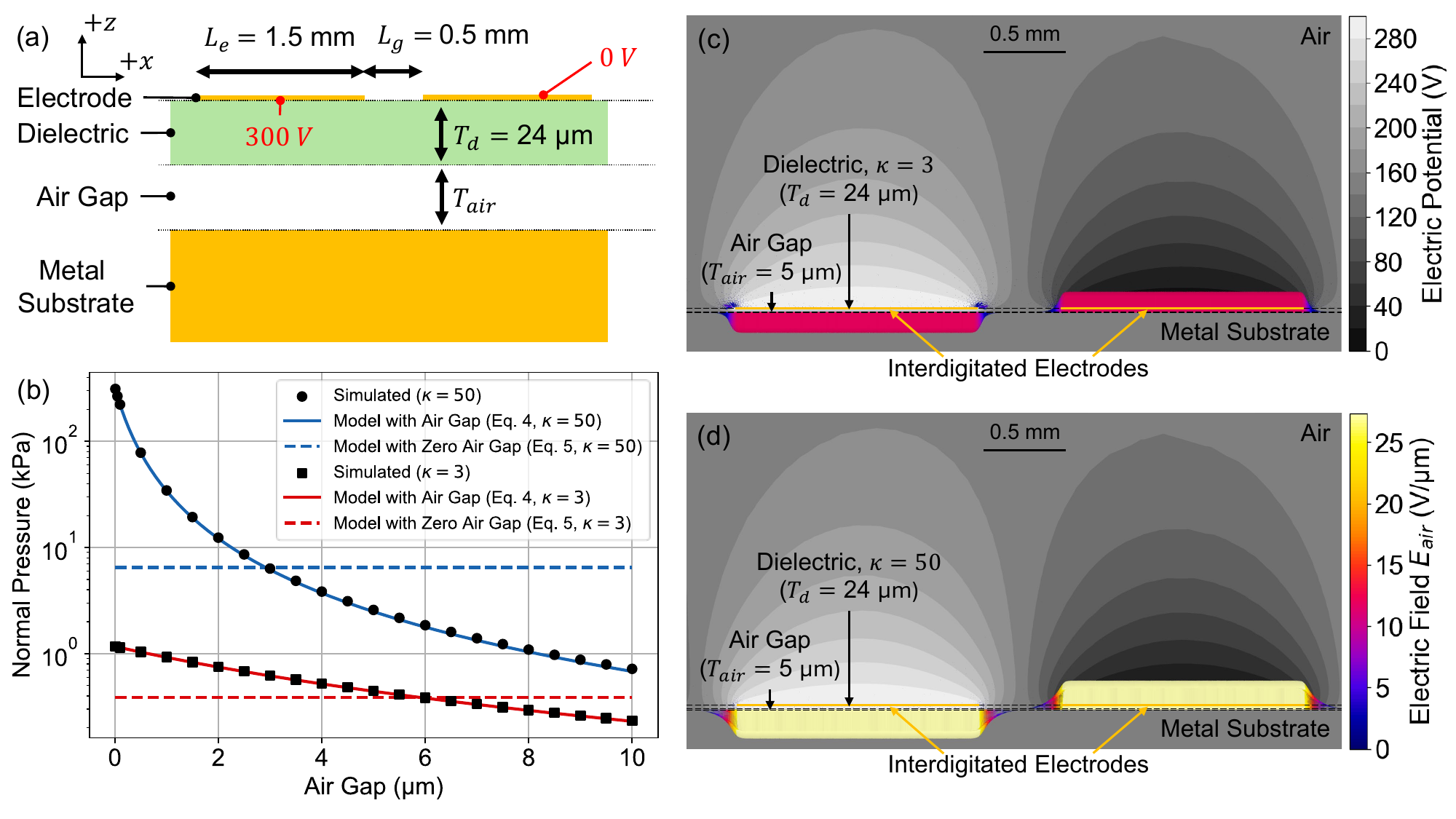}
	\caption{(a) Schematic of the geometry used for finite element simulation. (b) Comparison of the normal EA force predicted by a finite element simulation against the analytical models in Equations \ref{eq:electroadhesive_normal_force} and \ref{eq:force_electrostatic_noairgap} for dielectrics with relative permittivities $\kappa = 3$ and $\kappa = 50$. (c, d) Finite element analysis of the electric field (colored arrows) and potential (grayscale iso-contours) through an electroadhesive layer stack for a metallic substrate. The dielectric is a 24 \textmu m thick film dielectric with (c) $\kappa = 3$ and (d) $\kappa = 50$, and the air gap depicted here is 5 \textmu m thick. Both (c) and (d) use the same potential and electric field color scheme.}
	\label{fig:sim_force_vs_airgap}
\end{figure*}

While the air gap's importance is well-established empirically in the literature for designing high-force EA clutches \cite{AliAbbasi_Martinsen_Pettersen_Colgate_Basdogan_2024, Chen_Bergbreiter_2017}, Eq. \ref{eq:electroadhesive_normal_force} helps explain why. As depicted in Fig. \ref{fig:schematic_diagram}(a), the polarized dielectric's bound charge amplifies the air's electric field by an extra factor $\kappa$, enabling higher forces for very small air gaps. To verify this model, an electrostatics finite element simulation was developed in Altair Flux for an EA-ICE clutch with nominal dimensions (Table \ref{tab:nominal_dimensions}) and a 2 mm wide metal substrate. Fig. \ref{fig:sim_force_vs_airgap}(c, d) shows isovoltage contours and an electrical field heatmap through a cross-section of two interdigitated electrodes for dielectrics with either low ($\kappa = 3$) and high ($\kappa = 50$) relative permittivity. The metal substrate was simulated as a perfect conductor with floating potential, and the voltages were applied as boundary conditions on the dielectric's surface. Previous literature has primarily focused on finite element simulations of very closely spaced together electrodes and thus have had to account for significant fringing fields between adjacent fingers \cite{Cao_Sun_Fang_Qin_Yu_Feng_2016,Guo_Bamber_Chamberlain_Justham_Jackson_2016}. However, as shown in Fig. \ref{fig:sim_force_vs_airgap}(a), our larger electrode spacing ($L_g = L_e/3$) means that the electric field from each interdigitated finger affects the substrate almost independently of the electric field from adjacent fingers. Fig. \ref{fig:sim_force_vs_airgap}(b) compares the simulated normal pressure against Eq. \ref{eq:electroadhesive_normal_force} and Eq. \ref{eq:force_electrostatic_noairgap}, showing that Eq. \ref{eq:electroadhesive_normal_force} well approximates the finite element simulation to within 5\% even without a fringing field corrective factor. Fig. \ref{fig:sim_force_vs_airgap}(b) also shows that the air gap affects the electroadhesive normal pressure much more in systems with high-$\kappa$ dielectrics. For example, if $\kappa = 50$, even a few micrometers of displacement can result in orders of magnitude of force change.

\begin{table}[!t]
	\caption{Electroadhesive clutch and substrate dimensions, as shown in Fig. \ref{fig:schematic_diagram}(a, b)}
	\label{tab:nominal_dimensions}
	\centering
	\begin{tabular}{|c||c|}
		\hline
		$N$ & 28\\
		\hline
		$L_e$ (mm) & 1.5\\
		\hline
		$L_g$ (mm) & 0.5\\
		\hline
		$T_d$ (\textmu m) & 24\\
		\hline
		$L_s$ (mm) & 55.5\\
		\hline
		$w_s \times T_s$ (mm$\times$mm) & 2 $\times$ 2; 2.5 $\times$ 2.5; 3 $\times$ 3; 4 $\times$ 2; 5 $\times$ 2.5; 6 $\times$ 3 \\
		\hline
	\end{tabular}
\end{table}

While we leave a full retrospection to future work, we propose that factoring in the air gap via Eq. \ref{eq:electroadhesive_normal_force} can enable better predictions of EA clutch performance. For example, Hinchet and Shea \cite{Hinchet_Shea_2022} measured a capacitance of 0.76 nF cm$^{-2}$ for a parallel plate EA clutch at 150 V, which corresponds to an air gap of 610 nm (Eq. \ref{eq:t_air_from_capacitance}) and a predicted shear force capacity of 73.9 kPa (Eq. \ref{eq:electroadhesive_normal_force}). This is much closer to their measured 79.2 kPa than Eq. \ref{eq:force_electrostatic_noairgap}'s predicted 7.2 kPa. Similarly, Berdozzi et al. \cite{Berdozzi_Chen_Luzi_Fontana_Fassi_Molinari_Tosatti_Vertechy_2020} measured a capacitance change of 76.9 pF for an EA-ICE clutch at 1 kV, corresponding to an air gap of 9.2 \textmu m. Eq. \ref{eq:electroadhesive_normal_force} predicts a much closer shear force capacity of 2.54 kPa to their measured 2.02 kPa than Eq. \ref{eq:force_electrostatic_noairgap}'s predicted 15.3 kPa. Diller et al. \cite{Diller_Collins_Majidi_2018} also noted in their analysis that their data fits $\kappa^2$ much better than the $\kappa$ they were using from Eq. \ref{eq:force_electrostatic_noairgap}.

As a technical note, our derivation above only considers Coulombic electroadhesion, which affects both insulating and semiconducting dielectrics and which can create electrostatic attractive forces between both contacting and non-contacting surfaces. We do not extend our model to Johnsen-Rahbek electroadhesion, which primarily only affects semiconducting dielectrics in close contact with the substrate, because the dielectrics in this study are insulators ($\rho > 10^{10}$  $\Omega$ cm). We refer the reader to \cite{Nakamura_Yamamoto_2017, Persson_2021} for an in-depth comparison between Coulombic and Johnsen-Rahbek electroadhesion.

\subsection{Extending Quasistatic Analysis to the Transient Domain} \label{subsec:electrostatic_dynamics}
In this section, we extend the quasistatic analysis in Sec. \ref{subsec:electrostatic_force} to the transient domain, considering how polarization dynamics and the high voltage drive circuitry could affect EA dynamics. Contrary to assumptions in prior modeling work \cite{Chen_Liu_Wang_Zhu_Tao_Zhang_Jiang_2022, Nakamura_Yamamoto_2017}, we conclude that the upper limit of EA dynamics will be limited largely by mechanical rather than electrical dynamics.

\subsubsection{Dielectric Constant Dynamics} \label{subsec:dielectric_constanT_dynamics}
Sec. \ref{subsec:electrostatic_force}'s quasi-static analysis assumed that the substrate's induced charge instantaneously matches the deposited surface charge $\sigma$. This assumption is valid at our time scale for a metal substrate, in which charges can migrate in under a nanosecond. However, dielectrics move charges more slowly as polymer chains and interfacial charges reorient to be in line with the electric field \cite{Chen_Liu_Wang_Zhu_Tao_Zhang_Jiang_2022}. 

A dielectric's relative permittivity can be written as a function of angular frequency $\omega$ using the Cole-Cole equation:
\begin{equation} \label{eq:cole_cole}
	\kappa(\omega) = \kappa_\infty + \frac{\kappa_s - \kappa_\infty}{1 + (j\omega\tau)^\alpha}
\end{equation}
where $\kappa_\infty$ is the high-frequency limit of the relative permittivity, $\kappa_s$ is the low-frequency saturated relative permittivity, $\tau$ is the characteristic relaxation time, and $0 \leq \alpha \leq 1$ describes the broadness of the dielectric dispersion spectra. 

\begin{figure}[t]
	\centering
	\includegraphics[width=0.49\linewidth]{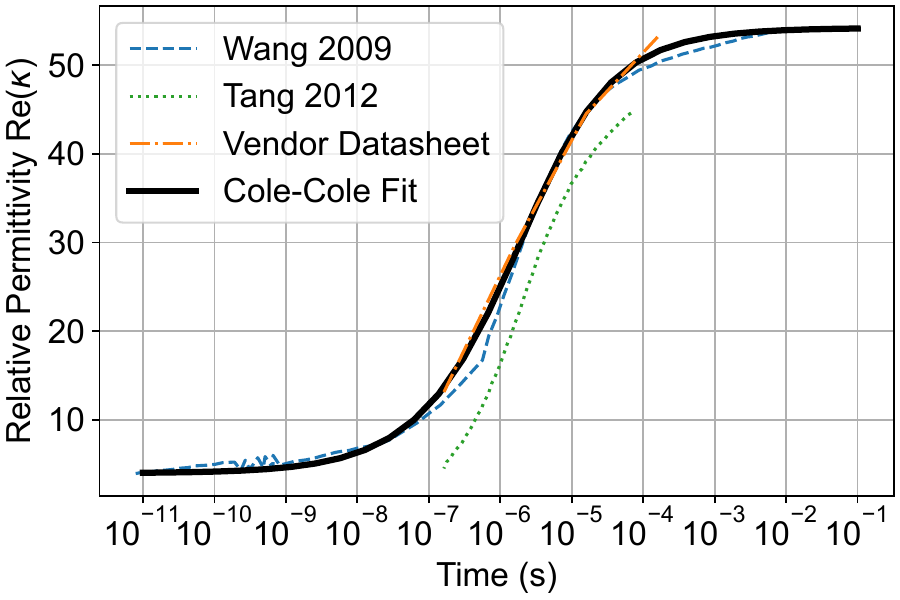}
	\caption{Step input response of a Cole-Cole dielectric constant model for P(VDF-TrFE-CFE). The model was least squares fit to experimental data from three sources, each measured across different frequency domains.}
	\label{fig:kappa_vs_frequency_colecole_fit}
\end{figure}

In this paper, we focus on the dielectric and relaxor ferroelectric P(VDF-TrFE-CFE) (poly(vinylidene fluoride-trifluoroethylene-chlorofluoroethylene)), which is commonly used in EA clutches in the literature for its high breakdown field and permittivity \cite{Rauf_Bernardo_Follmer_2023, Li_2021}. Modeling the charge as a step input, we can take the inverse Laplace transform:
\begin{equation} \label{eq:cole_cole_time}
	\kappa(t) = \kappa_\infty \delta(t) + (\kappa_s - \kappa_\infty) \left(\frac{t}{\tau}\right)^\alpha E_{\alpha, \alpha + 1}\left(- \left(\frac{t}{\tau}\right)^\alpha\right)
\end{equation}
where $\delta(t)$ is the Dirac delta function and $E_{\alpha, \beta}(z)$ is the Mittag-Leffler function. Fig. \ref{fig:kappa_vs_frequency_colecole_fit} shows the step input response to a fitted Cole-Cole model for P(VDF-TrFE-CFE) based on data from our vendor datasheet and from \cite{Wang_Lu_Lanagan_Zhang_2009, Tang_Lin_Sodano_2012}, which tested samples with a similar composition and annealing process to our own. Taking $\kappa_\infty = 4$ based on \cite{Wang_Lu_Lanagan_Zhang_2009}, the best fit coefficients were $\kappa_s = 54.2$, $\tau = 2.82$ \textmu s, and $\alpha = 0.562$. Interpreting the graph, after a charge $\sigma$ is deposited on (or removed from) an electrode, the dielectric can reach 93\% polarization (or depolarization) within the first 100 \textmu s. Similar curves can be drawn for other common EA dielectrics, and we conclude that while polarization dynamics set an important upper threshold on EA bandwidth, they are not primarily responsible for the 5-50 ms engagement times seen in prior literature (Fig. \ref{fig:schematic_diagram}(c)).

An important note is that the Cole-Cole model applies when the EA pad is driven by a power supply that can both source and sink charge. Prior modeling literature often asserts, based on the language for Maxwell-Wagner interfacial polarization, that the characteristic relaxation time should be on the order of seconds to minutes \cite{Chen_Liu_Wang_Zhu_Tao_Zhang_Jiang_2022, Nakamura_Yamamoto_2017, Chen_Zhang_Song_Fang_Sindersberger_Monkman_Guo_2020}. However, this assumes the EA pad is disconnected from a power supply and thus can only dissipate charge through internal resistance and stochastic depolarization. Although this is useful in applications such as perching robots or low-bandwidth tactile displays, where EA pads adhere to an insulating substrate using a power supply that intentionally does not sink the interfacial polarization charge \cite{Park_Drew_Follmer_Rivas-Davila_2020, AliAbbasi_Martinsen_Pettersen_Colgate_Basdogan_2024}, we proceed with the Cole-Cole model since most EA devices always stay connected to their power supply. Additional discussion and finite element analyses supporting a Cole-Cole polarization model instead of a Maxwell-Wagner polarization model for our application are presented in Supplementary Materials, Sec. S3.

\subsubsection{High Voltage Drive Circuit Dynamics} \label{subsec:high_voltage_drive_circuiT_dynamics}
The output resistance of the high voltage drive circuit is also important to consider when modeling EA dynamics. Output stages include common source amplifiers \cite{Rauf_Bernardo_Follmer_2023, Zhang_Gonzalez_Guo_Follmer_2019}, push-pull output stages such as half bridges \cite{Shultz_Harrison_2023} and full bridges \cite{Vechev_Hinchet_Coros_Thomaszewski_Hilliges_2022, Hinchet_Shea_2022, Hinchet_Shea_2020, Ramachandran_Shintake_Floreano_2019}, and optocoupler relays \cite{Diller_Collins_Majidi_2018, Han_Hajj-Ahmad_Cutkosky_2020}. Although the transistors in these circuits have response times far faster than EA's mechanical time constants, many small form factor DC-DC converters cannot handle the large current spikes that occur whenever these transistors flip on or off. Several previous papers thus use M$\Omega$-scale current-limiting resistors in series with their EA clutches, which we believe plays a large role in why they cite millisecond-scale rise and fall times for their high voltage drive circuits \cite{Rauf_Bernardo_Follmer_2023, Ramachandran_Shintake_Floreano_2019}. We expect that pairing a four-quadrant high voltage amplifier capable of handling these current spikes together with a fast output stage can enable better engagement and release times.

\subsection{Mechanical Dynamics} \label{subsec:mechanical_dynamics}
Until now, we have assumed the dielectric and substrate's surfaces are perfectly flat. However, as shown in Fig. \ref{fig:schematic_diagram}(a), the interface between an EA pad and substrate has many random asperities. As the asperities compress against each other, the apparent contact area between the two surfaces increases, resulting in what classical mechanics calls ``normal force'' \cite{Levine_Turner_Pikul_2021}.

We consider a simplified surface profile, where the average gap between the surfaces is $T_{air}$ and the distribution of asperities heights is Gaussian with standard deviation $\sigma_d$. Greenwood and Williamson's contact mechanics model \cite{Greenwood_Williamson_1966}, which is prevalent in tribology theory, models the stiffness force between a rigid substrate and deformable surface as:
\begin{equation} \label{eq:force_displacement_asperity}
	F_k(T_{air}) = kL_sw_s\sigma_d^{1.5}G_{1.5}\left(\frac{T_{air}}{\sigma_d}\right)
\end{equation}
where $k$ is a fitted parameter related to the asperity geometry and material parameters. $G_{1.5}(h)$ is a function with the form:
\begin{equation} \label{eq:GW_model_G32}
	G_{1.5}(h) = \frac{\sqrt{h}e^{-\frac{h^2}{4}}}{4\sqrt{\pi}}\left[(h^2 + 1)K_{\frac{1}{4}}\left(\frac{h^2}{4}\right) - h^2K_{\frac{3}{4}}\left(\frac{h^2}{4}\right)\right]
\end{equation}
where $K_\nu(z)$ is the Basset function. $G_{1.5}(\frac{T_{air}}{\sigma_d})\approx 0.43$ when $T_{air} << \sigma_d$, and quickly drops off to zero once $T_{air} > \sigma_d$.

Because of the large area and small gap, the film will likely experience squeeze-film damping as its predominant form of viscous damping. Griffin et al. \cite{Griffin_Richardson_Yamanami_1966} derived a first-order approximation for large flat plates:
\begin{equation} \label{eq:damping_force}
	F_b(T_{air}) = -\left( \frac{96\mu_{air} \min(L_s, w_s)^3\max(L_s, w_s)}{\pi^4 T_{air}^3} \right) \dot{T}_{air}
\end{equation} 
where $\mu_{air} =$ 1.85 $\cdot 10^{-5}$ N s m$^{-2}$ is air's dynamic viscosity. To account for the Gaussian distribution of asperity heights, we can find the average damping force by integrating $F_b(T_{air})$ with the asperity height's probability density function \cite{Persson_2021, Ayyildiz_Scaraggi_Sirin_Basdogan_Persson_2018}:
\begin{equation} \label{eq:F_b_gaussian}
	F_b'(T_{air}) = \int_0^\infty F_b(\tau_{air})\left(\frac{e^{-(T_{air} - \tau_{air})/2\sigma_d^2}}{\sigma_d \sqrt{2\pi}}  \right) \textrm{d}\tau_{air}
\end{equation}

Conveniently, the average gap between surfaces used in contact mechanics models is the same $T_{air}$ used in our EA normal force equation $F_{ea}(T_{air})$ (Eq. \ref{eq:electroadhesive_normal_force}). Thus, to account for the effect of surface roughness, we can use a similar integral over the asperity height probability density function to define the average EA normal force $F_{ea}'$:
\begin{equation} \label{eq:F_ea_gaussian}
	F_{ea}'(T_{air}) = \int_{-\frac{T_d}{\kappa}}^\infty F_{ea}(\tau_{air})\left(\frac{e^{-(T_{air} - \tau_{air})/2\sigma_d^2}}{\sigma_d \sqrt{2\pi}}  \right) \textrm{d}\tau_{air}
\end{equation}
where the negative integral range accounts for when the dielectric compresses against asperities. These alterations have very little effect when $T_{air} \gg \sigma_d$, but for our nominal $T_d$ it can increase $F_{ea}'$ by up to 11.1$\times$ when $T_{air} \approx \sigma_d$. This formulation only accounts for EA force in the $z$ direction and assumes the influence of fringing fields between asperities is negligible, similar to prior work \cite{Persson_2021, Ayyildiz_Scaraggi_Sirin_Basdogan_Persson_2018}, although future work should further investigate the validity of this assumption.

We also explored a similar averaging formulation of the parallel plate capacitance equation in Eq. \ref{eq:capacitance}, but we found it made very little difference for our system. This matches the findings of prior work on capacitive contact profilometers \cite{Garbini_Koh_Jorgensen_Ramulu_1992, Nowicki_Jarkiewicz_1998, Lanyi_1998, Guadarrama-Santana_Garcia-Valenzuela_Bruce_Hernandez-Cordero_2003}, which generally find that micrometer-scale surface roughnesses only alter the ideal flat plate capacitance by a few percent over centimeter-scale overlap areas so long as $T_{air} > \sigma_d$. Using similar reasoning, prior electroadhesion studies have primarily used parallel plate capacitance models to estimate their systems' average air gap \cite{Shultz_Peshkin_Colgate_2018, AliAbbasi_Martinsen_Pettersen_Colgate_Basdogan_2024, Wang_Zhang_Li_Hong_Wang_2019, Xie_Wang_Zhao_Wang_2019}.

\subsection{Full Electromechanical Dynamics Model} \label{subsec:electromechanical_dynamics_model}
\begin{figure}[t]
	\centering
	\includegraphics[width=0.49\linewidth]{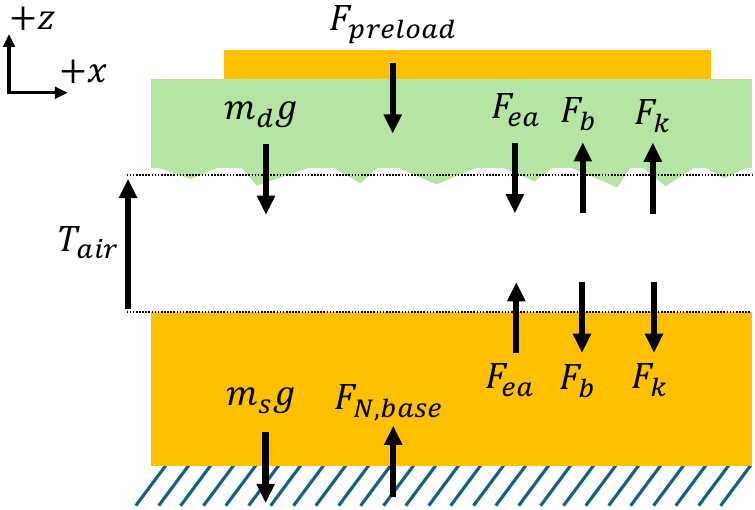}
	\caption{Free body diagram of forces on the dielectric and substrate.}
	\label{fig:free_body_diagram}
\end{figure}

Combining the equations from the last three sections together, we can write equations of motion in the $z$ axis for the dielectric and substrate. We assume the substrate is much heavier than the dielectric (i.e., fixed in $z$-height), while the dielectric is free to move in the $z$ direction. The dielectric also has a small preload force $F_{preload}$ applied to help prevent wrinkles and ensure even contact between the surfaces, in line with prior experimental work \cite{Chen_Bergbreiter_2017, Yamamoto_Nakashima_Higuchi_2007, Schaller_Shea_2023, Guo_Tailor_Bamber_Chamberlain_Justham_Jackson_2015}. Referencing the free body diagram in Fig. \ref{fig:free_body_diagram}, the dielectric's equation of motion along the $z$ axis is:
\begin{equation} \label{eq:eom_dielectric_z}
	m_d\ddot{T}_{air} = F_k + F_b' - F_{ea}' - m_dg - F_{preload}
\end{equation}
where the first term on the right side comes from contact mechanics (Eq. \ref{eq:force_displacement_asperity}), the second term comes from damping (Eq. \ref{eq:F_b_gaussian}), the third comes from electrostatics (Eq. \ref{eq:F_ea_gaussian}, but incorporating the rise and fall times of $\kappa$ and $V$ as described in Sec. \ref{subsec:electrostatic_dynamics}), and the fourth comes from gravity on the dielectric mass $m_d$.

The substrate's equation of motion along the $z$ axis is:
\begin{equation} \label{eq:eom_substrate_z}
	0 = F_{N,base} + F_{ea}' - F_k - F_b' - m_sg
\end{equation}
where $F_{N,base}$ is the normal force from the base on the substrate and $m_s$ is the mass of the substrate. Since $F_{N,base}$ does not directly affect the dynamics, we leave it as is instead of modeling its contact mechanics and air gap. The system of equations given by Eqs. \ref{eq:eom_dielectric_z} and \ref{eq:eom_substrate_z} can be solved as an initial value problem with the state $[T_{air}, \dot{T}_{air}]$. 

We denote the substrate's coefficients of friction with the dielectric and base as $\mu_d$ and $\mu_{base}$, respectively. After solving Eqs. \ref{eq:eom_dielectric_z} and \ref{eq:eom_substrate_z} for $T_{air}$, we can predict the shear friction on the substrate while the substrate is moving and while the dielectric and base remain fixed along the substrate's axis of motion:
\begin{equation} \label{eq:shear_force}
	F_{shear} = \mu_{base}F_{N,base} + \mu_dF_k
\end{equation}

Once the gap settles ($\ddot{T}_{air} = 0$) and after solving Eq. \ref{eq:eom_dielectric_z} for $T_{air}$, we can predict the EA clutch's shear force capacity:
\begin{equation} \label{eq:shear_force_capacity}
	F_{shear,max} = \mu_{base}(m_dg + m_sg + F_{preload}) + \mu_dF_k
\end{equation}

The preload force can also be estimated by measuring the force $F_{shear, V=0}$ required to push the substrate when $V = 0$:
\begin{equation} \label{eq:F_preload}
	F_{preload} = \frac{F_{shear, V=0} - \mu_{base}m_sg - \mu_dm_dg}{\mu_d + \mu_{base}}
\end{equation}

\section{Shear Force Capacity Results} \label{sec:experimental_measurements_shear_force_capacity}

\subsection{Experimental Setup and Testing Procedure} \label{subsec:fts_experimental_test_setup}
\begin{figure}[!t]
	\includegraphics[width=\linewidth]{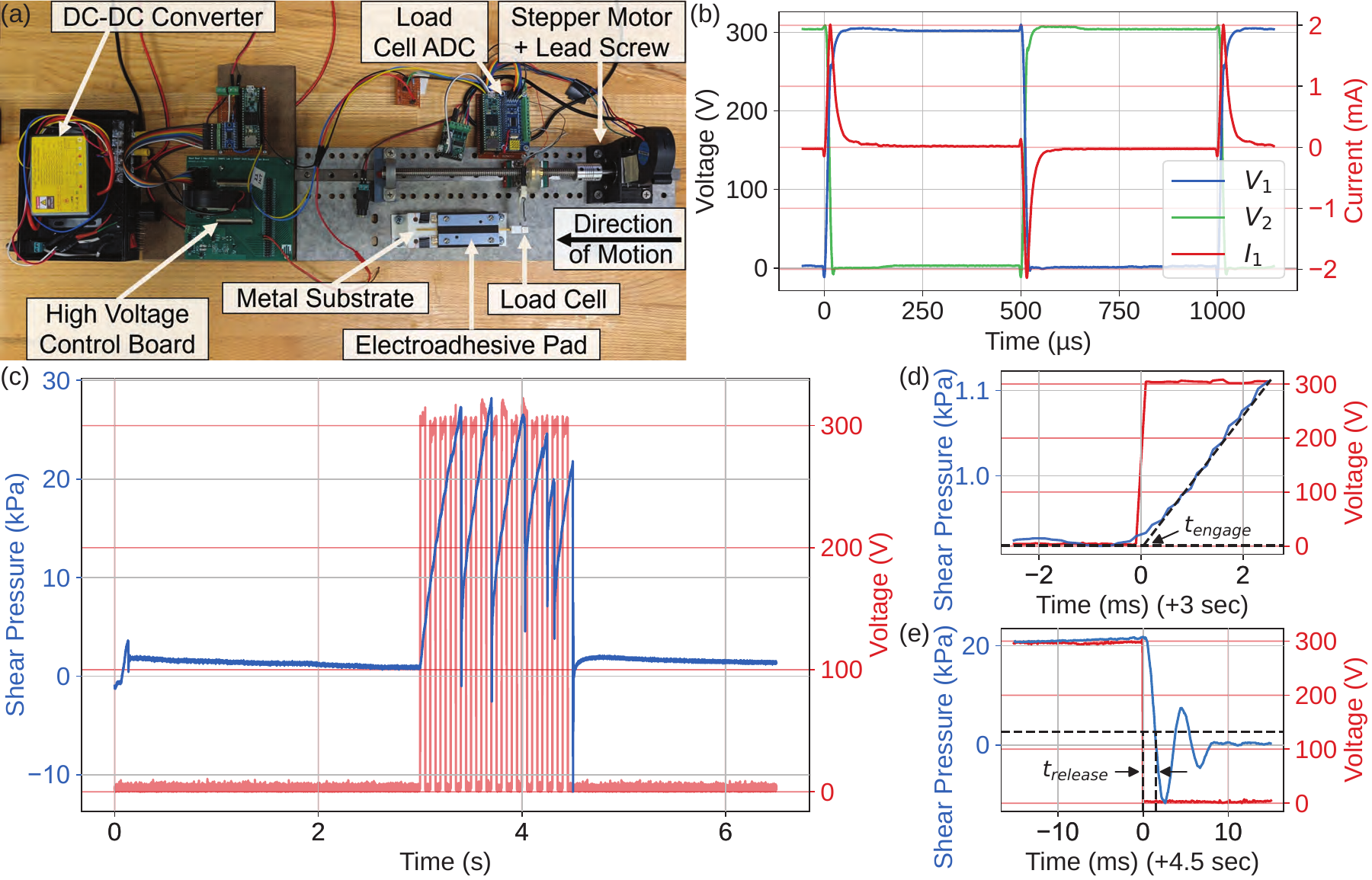}
	\caption{(a) Experimental test setup for measuring EA shear force as the load cell on a motorized stage pushes against the substrate. (b) Sample voltage and current traces for a 300 V, 1 kHz bipolar square wave input, measured across a 100 k$\Omega$ resistor in series with an EA clutch. (c) Load cell force measured over the course of a shear force capacity test. The drive signal, shown for one electrode, was a 300 V, 10 Hz bipolar square wave. (d) Calculation procedure for the engagement time of this sample. The force profile's rounded corner when the voltage rises is due to filtering, so the engagement time is measured when the force profile after the voltage rises linearly interpolates to the average force before the voltage rises (shown via dotted black lines). (e) Calculation procedure for the release time of this sample, measured as the time between when the voltage falls and when the load cell falls to within 10\% of its final settling value. Using these metrics, for the sample shown $t_{engage}$ = 2.6 \textmu s and $t_{release} = $ 1.49 ms.}
	\label{fig:force_test_setup}
\end{figure}

To test our model, we built the experimental test setup in Fig. \ref{fig:force_test_setup}(a) to measure the shear force capacity, engagement time, and release time of an EA pad to a metallic substrate across a variety of geometric and operating parameters. Similar to prior work on both parallel plate EA clutches \cite{Diller_Collins_Majidi_2018, Hinchet_Shea_2020} and EA-ICE clutches \cite{Zhang_Gonzalez_Guo_Follmer_2019, Berdozzi_Chen_Luzi_Fontana_Fassi_Molinari_Tosatti_Vertechy_2020, Kim_Lee_Bhuyan_Wei_Kim_Shimizu_Shintake_Park_2023}, we created a tensile testing machine by mounting a 500g load cell (UXCell, China) on a motorized linear slide. The stepper motor moving the slide block was driven at 1/256 microstepping, and the load cell was sampled at 38.4 kHz (ADS1263, Texas Instruments, USA). 

The EA pad was made using a 24 \textmu m thick P(VDF-TrFE-CFE) film with 7 mol\% CFE and a sputtered gold layer on one side (PolyK Technologies, LLC, USA). An interdigitated electrode pattern with dimensions in Table \ref{tab:nominal_dimensions} was etched into the gold using a UV laser cutter (Series 3500, DPSS Lasers Inc., USA). Enameled copper wire was adhered to each electrode using silver conductive epoxy. As shown in Fig. \ref{fig:schematic_diagram}(b), the interdigitated electrodes were designed to be wider than the substrate to mimic prior work on using EA clutches to manipulate small metallic objects \cite{Zhang_Gonzalez_Guo_Follmer_2019}, and to mitigate inconsistencies if the pin slipped sideways when reset between runs. The high voltage drive signal was powered by a 50 mA rated DC-DC converter (AHV12V500V50MAW, Analog Technologies, USA) and a high voltage shift register with a half-bridge output stage (HV507, Microchip Technology Inc., USA). 

We used six brass substrates (Albion Alloys, England) with different dimensions shown in Table \ref{tab:nominal_dimensions}. Screws were tightened through mounting points in the assembly to lightly press the dielectric into the substrate, which helped reduce film wrinkling under high shear stress. Pressure was evenly distributed along the sides using 3D printed brackets and a 1/16'' closed-cell blended EPDM foam (McMaster-Carr Supply Company, USA). The foam was chosen for its low modulus (comparable to 5A durometer silicone) and low tackiness. This setup constrains the brass substrate to move primarily in the $x$ direction and constrains the dielectric film to move primarily in the $z$ direction, in line with our model assumptions in Sec. \ref{sec:modeling_electroadhesion}. The boundary constraints also apply a small preload force $F_{preload} < 0.25$ N to our dielectric film, which is among the lower range of preload forces used in prior work \cite{Zhang_Gonzalez_Guo_Follmer_2019, Shultz_Peshkin_Colgate_2018, Nakamura_Yamamoto_2017, AliAbbasi_Martinsen_Pettersen_Colgate_Basdogan_2024, Guo_Tailor_Bamber_Chamberlain_Justham_Jackson_2015}. Almost all prior works in Fig. \ref{fig:schematic_diagram}(b) apply a preload force to ensure conformality and prevent dielectric film wrinkling. Although smaller preload forces tend to be better for dynamics measurements to avoid additional sources of noise, in Sec. \ref{sec:experimental_measurements_engagement_release}, we explore adding extra clamping pressure to test the effect of higher preload forces within the range seen in prior work.

The base that the substrate moves on was 3D printed with a glossy finish from VeroWhitePlus photopolymer using a Strasys Objet24 V3. To measure the coefficient of friction, we took the ratio between tangential and normal force across 10 trials, measuring $\mu_{base}^{static} = 0.281$ and $\mu_{base}^{kinetic} = 0.173$ between the brass substrate and the base plate; and $\mu_d^{static} = 0.188$ and $\mu_d^{kinetic} = 0.154$ between the substrate and the dielectric film. Additional testing setup implementation details are included in Supplementary Materials, Sec. S4.

Fig. \ref{fig:force_test_setup}(b) shows a sample voltage and current profile for a 300 V, 1 kHz bipolar square wave drive signal. Current was measured by using a differential oscilloscope probe (DP10013, Shenzhen Micsig Technology Co., Ltd., China) across a 100 k$\Omega$ resistor placed in series with the clutch. The voltage was sampled on a Teensy 4.1 microcontroller (PJRC, USA) using a 100:1 resistive divider (10.1 M$\Omega$ total) placed in parallel with the clutch. We observed rise times (10\% - 90\%) of 8.3 \textmu s and fall times of 5.2 \textmu s, which assuming an ideal exponential RC curve correspond to time constants $\tau_{RC}\approx$ 2.4 - 3.8 \textmu s. These RC constants make sense; as shown in Fig. \ref{fig:capacitance_gap_vs_weight}(b), we expect our EA pad to have capacitances from 20 - 40 pF when adhering to a 2 mm substrate, which corresponds to a predicted RC time constant of 2 - 4 \textmu s over the $\approx$100 k$\Omega$ effective parallel resistance (the 100 k$\Omega$ series current sense resistor in parallel with the 10.1 M$\Omega$ parallel voltage sense resistor). The average leakage current from the power supply after the 300 V drive voltage saturates is about 26.1 \textmu A, which is accounted for by our 10.2 M$\Omega$ effective series resistance. The lack of additional current draw further validates our choice of a Coulombic electroadhesion force model instead of a Johnsen-Rahbek model, as we discussed in Sec. \ref{subsec:electrostatic_force}. Multiplying the voltage and current traces for both electrodes gives an average power consumption of 1.43 mW. During the rest of our experimental testing, we kept the 10.1 M$\Omega$ parallel voltage sense resistor but removed the 100 k$\Omega$ series current sense resistor to further improve our high voltage drive circuit's rise and fall times.

Fig. \ref{fig:force_test_setup}(c) shows a sample experimental run. We include the raw data for 10 additional experimental runs in Supplementary Materials, Sec. S5 and a video of a run in Supplementary Video 1. The stepper motor drives the substrate forward at 0.5 mm/s for 3 seconds to establish a baseline kinetic friction force reading. We then turned on the voltage while continuing to drive the motor for 1.5 seconds. Finally, the motor is stopped, the voltage is removed, and the load cell is allowed to settle for 2 seconds. We chose to push the rigid substrate instead of pulling on our soft dielectric film to avoid plastic deformation, in line with similar guidance by the ISO 8295 standard on measuring the frictional properties of thin films \cite{ISO8295}. A timing diagram of the actuation and sensing signals is shown in Supplementary Materials, Sec. S4.1. The ADC is synced using the Teensy microcontroller to always sample the load cell and high voltage drive signal after the stepper motor settles its position, and the first reading is timed to occur about 4 \textmu s after the high voltage first turns on.

\subsection{Fitting Contact Mechanics Model} \label{subsec:fitting_contact_mechanics_model}
\begin{figure}[t]
	\centering
	\includegraphics[width=\columnwidth]{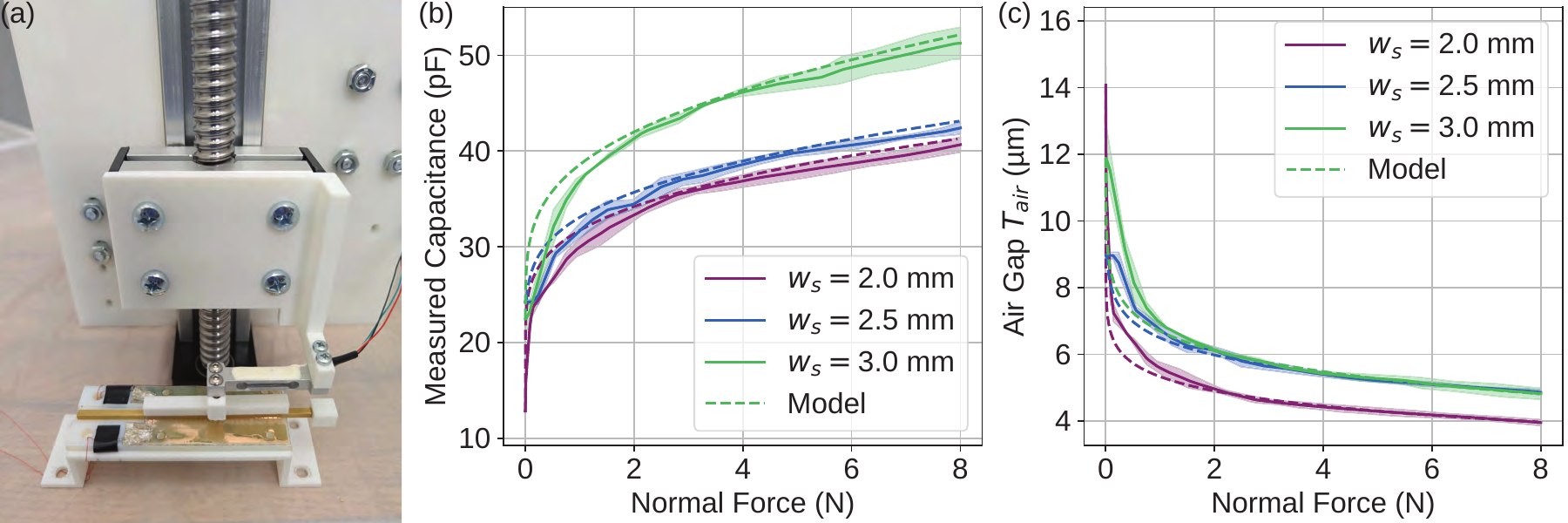}
	\caption{(a) Experimental test setup for measuring the EA pad capacitance over a range of applied normal forces. Given the measured capacitance values shown in (b), Eq. \ref{eq:t_air_from_capacitance} was used to estimate the air gaps as shown in (c). The average and standard deviation across 3 trials are shown. The contact mechanics model in Eq. \ref{eq:force_displacement_asperity} was least squares fit to the curves in (c). Using these fitted parameters, the model's predicted capacitances across the normal force range are also plotted using dashed lines in (b).}
	\label{fig:capacitance_gap_vs_weight}
\end{figure}

To fit the parameters $k$ and $\sigma_d$ in Eq. \ref{eq:force_displacement_asperity}, we measured the capacitance of our EA pad against brass substrates of different widths across a range of applied normal forces, as shown in Fig. \ref{fig:capacitance_gap_vs_weight}. The EA pad's capacitance was read using a Keysight U1701B capacitance meter, after subtracting parasitic capacitance. The air gap was then estimated using Eq. \ref{eq:t_air_from_capacitance}. Fitting the air gap and normal force data in Fig. \ref{fig:capacitance_gap_vs_weight}(c) to Eq. \ref{eq:force_displacement_asperity}, our best fit parameters were $k = 5.39 \cdot 10^{14}$ N m$^{-3.5}$ and $\sigma_d = 2.80$ \textmu m. This calibration procedure is inspired by prior work on capacitive contact profilometry \cite{Garbini_Koh_Jorgensen_Ramulu_1992, Nowicki_Jarkiewicz_1998, Lanyi_1998, Guadarrama-Santana_Garcia-Valenzuela_Bruce_Hernandez-Cordero_2003} and presents an accessible alternative to traditional optical or stylus-based surface profilometry for characterizing EA dielectric surfaces \cite{Schaller_Shea_2023, Kim_Lee_Bhuyan_Wei_Kim_Shimizu_Shintake_Park_2023}.

As a technical note, we also tried fitting our capacitance data in Fig. \ref{fig:capacitance_gap_vs_weight}(b) to the surface roughness-averaged capacitance using a similar integral formulation as Eq. \ref{eq:F_ea_gaussian}. However, our best fit parameters were almost identical to the ones obtained by our parallel plate capacitance model in Eq. \ref{eq:capacitance} and Eq. \ref{eq:t_air_from_capacitance}. This aligns with expectations based on prior work applying capacitive contact profilometry to surfaces with micrometer-scale surface roughnesses \cite{Garbini_Koh_Jorgensen_Ramulu_1992, Nowicki_Jarkiewicz_1998, Lanyi_1998, Guadarrama-Santana_Garcia-Valenzuela_Bruce_Hernandez-Cordero_2003}. It also matches the methods used by prior EA studies to estimate their clutches' average air gap \cite{Shultz_Peshkin_Colgate_2018, AliAbbasi_Martinsen_Pettersen_Colgate_Basdogan_2024, Wang_Zhang_Li_Hong_Wang_2019, Xie_Wang_Zhao_Wang_2019}.

\subsection{Shear Force Capacity Measurements and Modeling} \label{subsec:fts_shear_force_capacity}
\begin{figure}[!t]
	\includegraphics[width=0.49\columnwidth]{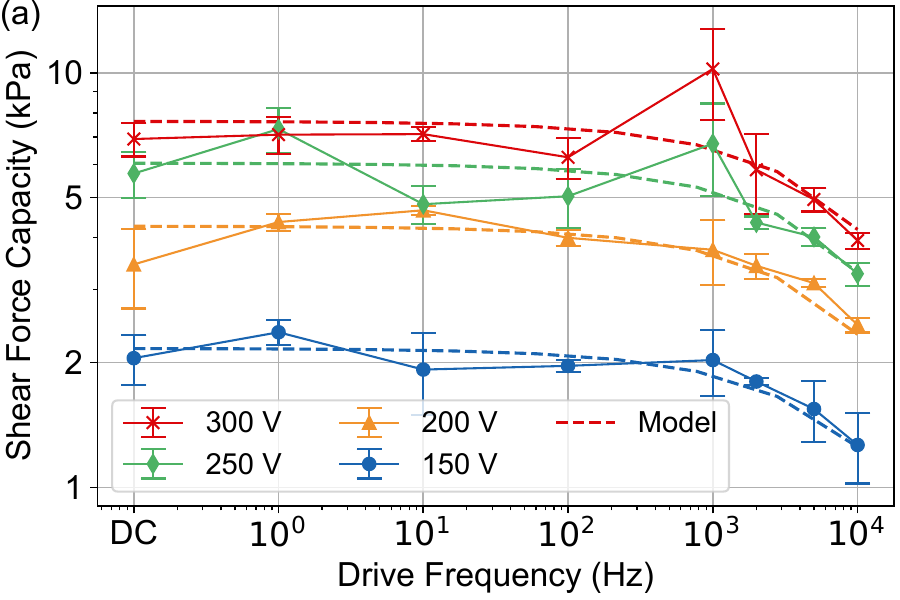}
	\includegraphics[width=0.49\columnwidth]{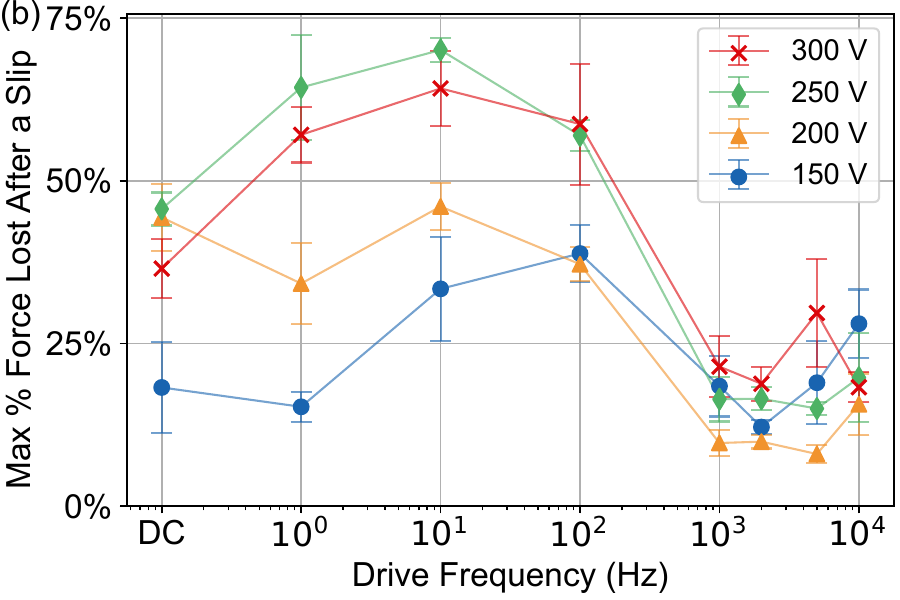}
	\caption{(a) Shear force capacity and (b) max slip percentage measured for a 2 mm wide substrate at different voltages and drive frequencies, averaged over at least 5 trials per condition.}
	\label{fig:force_vs_voltage_vs_frequency}
\end{figure}

Although maximizing shear force capacity is not a primary objective of this paper, understanding how different parameters affect our clutch's force capacity can help validate whether our model is physically sound. Fig. \ref{fig:force_vs_voltage_vs_frequency}(a) shows the measured and predicted (Eq. \ref{eq:shear_force_capacity}) shear force capacities at different voltages and bipolar square wave AC drive frequencies. The order of voltages and frequencies tested was randomized.

When applying our model, we found that our analytical EA force equation was slightly inconsistent from what was needed to produce the measured shear capacities, so we fitted a multiplier $\lambda^{fit}_{ea}(V)$ for $F_{ea}$ to each voltage's curve in Fig. \ref{fig:force_vs_voltage_vs_frequency}(a):
\begin{equation}
	F_{ea,adj}' = \lambda^{fit}_{ea}(V)F_{ea}'
\end{equation}

We found that the best fit values followed a roughly linear relationship $\lambda^{fit}_{ea}(V) = 2.40 - 0.0058V$ ($R^2=0.98$).
The average value of $\lambda^{fit}_{ea}$ over all samples was $1.08$ $(\sigma = 0.33)$. For context, although fitted multipliers are almost universal in prior work to account for hard-to-model physical and electrical non-idealities \cite{Guo_Tailor_Bamber_Chamberlain_Justham_Jackson_2015}, they tend to have a wide range of magnitudes ranging anywhere from $10^{-4}$ \cite{Feizi_Atashzar_Kermani_Patel_2022} to 10 \cite{Diller_Collins_Majidi_2018}. Fitted voltage-dependent multipliers have seen particular success in recent studies \cite{Levine_Turner_Pikul_2021, Basdogan_Sormoli_Sirin_2020, Kanno_Kato_Yoshioka_Nishio_Tsubone_2006, Levine_Lee_Campbell_McBride_Kim_Turner_Hayward_Pikul_2023}, drawing inspiration from the fact that many parasitics and dielectric behaviors like electrostriction are voltage-dependent. Because our average $\lambda^{fit}_{ea}$ was about 1 and because the model accurately predicted the shear pressure capacity's decay at high frequencies, we take these results as confirmation that our electromechanical model is, for the most part, physically sound. We leave in-depth modeling of these voltage-dependent behaviors to future work, and for the purposes of this study we proceed with the fitted multiplier $\lambda_{ea}^{fit}(V)$.

The -3 dB bandwidth for all voltages is experimentally over 10 kHz, and is predicted by our model to be about 13.3 kHz. For comparison, Chen et al. \cite{Chen_Liu_Wang_Zhu_Tao_Zhang_Jiang_2022} measured a -3 dB frequency of about 150 Hz for both square and sine AC drive signals using a glass substrate, Zhang et al. \cite{Zhang_Gonzalez_Guo_Follmer_2019} measured a -3 dB frequency of about 130 Hz for sine AC drive signals against a brass metal substrate, and Shultz et al. \cite{Shultz_Peshkin_Colgate_2018} measured a -3 dB frequency of around 75 Hz for sine AC drive signals using a human fingertip as the substrate. We predict the difference is due largely to the improved dynamics of our drive circuitry and our use of a metal substrate. Additional comparisons between our experimental results and prior literature are provided in Supplementary Materials, Sec. S6.

\subsection{Force Slip vs. Frequency}
As shown in the sample force profile in Fig. \ref{fig:force_test_setup}(c), EA clutches lose a significant amount of their load-carrying capacity after a slip, with similar clutches in the literature often losing 50-70\% of their measured force and requiring tens of milliseconds before the clutch re-engages \cite{Diller_Collins_Majidi_2018, Hinchet_Shea_2020, Chen_Bergbreiter_2017}. This failure mode is a major deterrent to using EA clutches in safety-critical applications. Fig. \ref{fig:force_vs_voltage_vs_frequency}(b) shows how the worst slip in each of our test runs compares across different voltages and frequencies, measured as the percentage of force lost after the slip. We found that the slip percentage decreases drastically at higher frequencies, with 250 V seeing a 4.7$\times$ reduction on average between the 10 Hz and 5 kHz drive frequencies. Our numerical values cannot be compared directly with prior work, since the slip percentage also varies with load cell dynamics, but prior work has primarily used DC or low-frequency ($\leq$10 Hz) AC drive signals. Our hypothesis, which we explore further in Sec. \ref{sec:experimental_measurements_engagement_release}, is that higher drive frequencies should reduce space charge buildup in the dielectric, enabling smoother release and faster re-engagement.

\section{Engagement and Release Time Results} \label{sec:experimental_measurements_engagement_release}
Both faster engagement times and faster release times are important to improving the effective bandwidth of EA clutches. We focus first on the release times of our EA-ICE clutches because, as shown in Fig. \ref{fig:schematic_diagram}(c), they tend to be much larger than the engagement times and thus a much larger component of EA bandwidth.

\subsection{Release Time Testing and Simulation Methods} \label{subsec:fts_release_time}

As shown in Fig. \ref{fig:force_test_setup}(e) and consistent with previous work \cite{Diller_Collins_Majidi_2018, Hinchet_Shea_2020}, we measured the release time for our EA clutches as the time between when the voltage begins to fall and when the load cell falls 90\% of the way to its final steady-state value. 

To simulate the release dynamics, in addition to the coupled equations of motion Eqs. \ref{eq:eom_dielectric_z} and \ref{eq:eom_substrate_z} for $T_{air}$, we also modeled the load cell dynamics along the substrate's axis of motion:
\begin{equation}
	m_{lc}\ddot{x}_{lc} = -k_{lc}x_{lc} - b_{lc}\dot{x}_{lc} - F_{shear}\textrm{sign}(\dot{x}_{lc})
\end{equation}
where $F_{shear}$, given by Eq. \ref{eq:shear_force}, is calculated using static friction coefficients when $\dot{x}_{lc}=0$ and kinetic friction coefficients otherwise. We measured the load cell's spring constant ($k_{lc} = 18.0$ N mm$^{-1}$), damping constant ($b_{lc} = 0.43$ kg s$^{-1}$), resonance frequency ($\omega_{lc} = 643.4$ Hz), and effective mass ($m_{lc} = 1.10$ g) by recording its response to step and ramp inputs to the load cell's position by the stepper motor. While the load cell dynamics were important to include in the model to fit the data, the load cell's quarter-period $\omega_{lc}/4 = 388.6$ \textmu s is small enough that we can expect the clutch's electromechanical dynamics to be the primary contributor to the measured release time. Additional discussion motivating this conclusion is presented in Supplementary Materials, Sec. S7.

For each experimental run, we estimated $F_{preload}$ by measuring the kinetic friction on the substrate during the first 3 sec before the voltage is enabled and applying Eq. \ref{eq:F_preload}. We also measured the ratio $\frac{F_{shear}(t_r)}{F_{shear,max}}$ between the load cell's applied force right before the voltage was released and the load cell's maximum measured force. Using $F_{preload}$, we solved the coupled equations of motion Eqs. \ref{eq:eom_dielectric_z} and \ref{eq:eom_substrate_z} to calculate $T_{air}(t_r)$ at the moment of release. We then estimated the shear force capacity $F_{shear,max}^{predicted}$ of the EA pad using Eq. \ref{eq:shear_force_capacity}. We estimated the load cell's initial position as:
\begin{equation} \label{eq:release_sim_initial_condition}
	x_{lc}(t_r) = \left(\frac{F_{shear,max}^{predicted} }{k_{lc}}\right)\left(\frac{F_{shear}(t_r)}{F_{shear,max}}\right)
\end{equation}
where the second term on the right side is the value for each run, representing how far the load cell has stretched relative to its maximum displacement before the voltage released. The initial condition for the simulation is taken at the estimated positions $T_{air}(t_r)$ and $x_{lc}(t_r)$ and with zero initial velocity. The simulation ends when the load cell returns to $x_{lc} = 0$.

\begin{figure}[!t]
	\centering
	\includegraphics[width=0.49\linewidth]{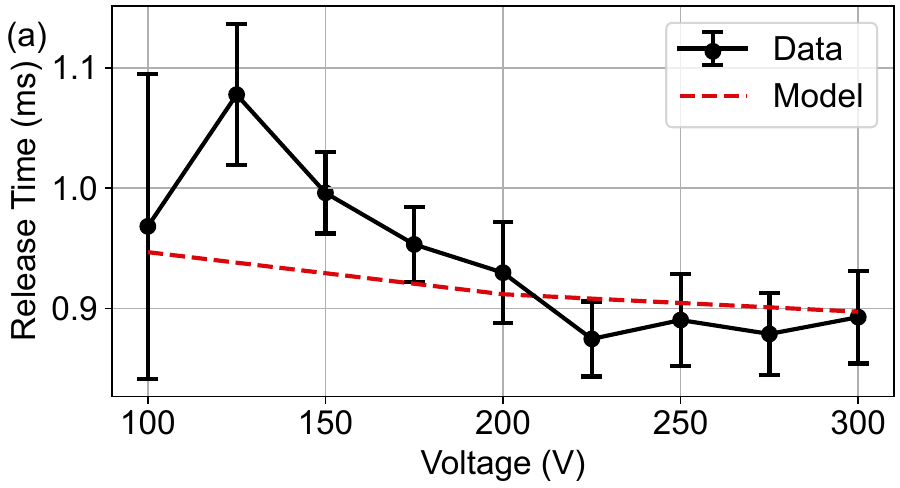}
	\includegraphics[width=0.49\linewidth]{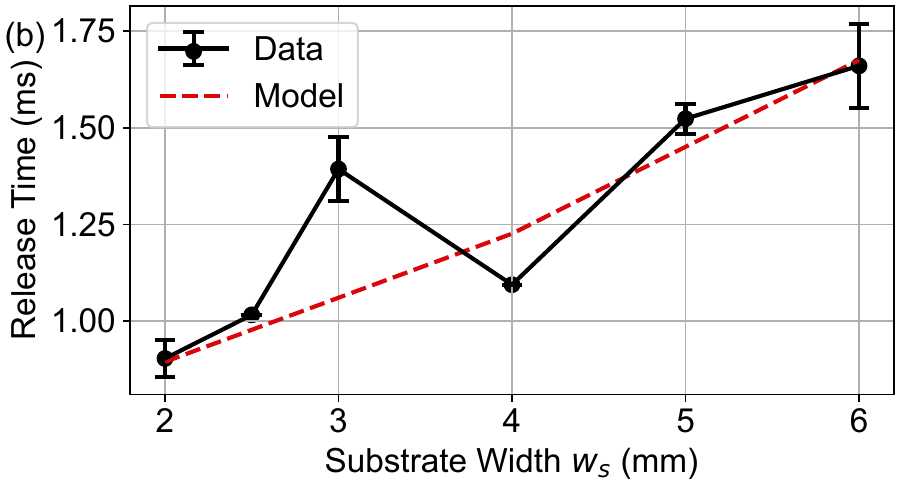}
	\includegraphics[width=0.49\linewidth]{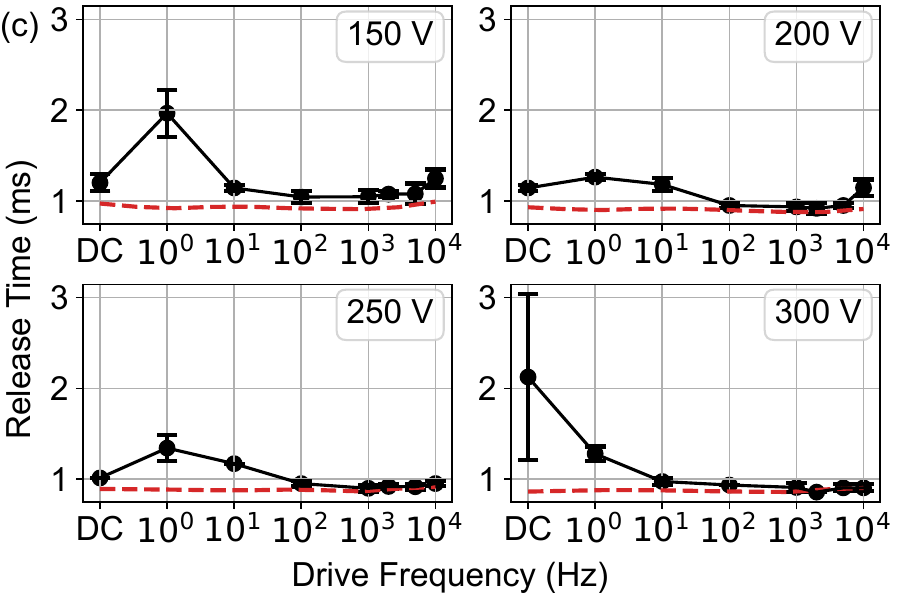}
	\includegraphics[width=0.49\linewidth]{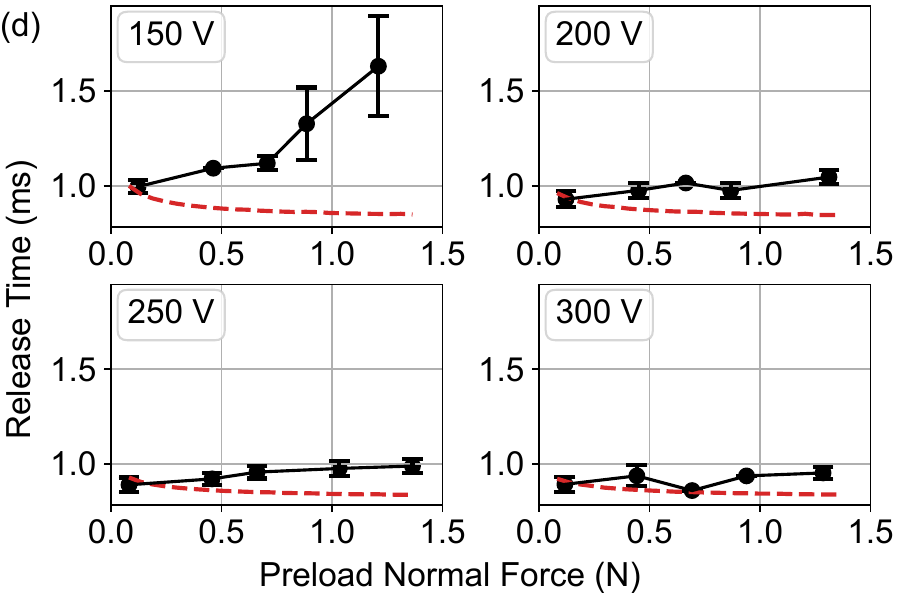}
	\caption{Release time averaged over at least 4 trials per condition for different (a) voltages, (b) substrate widths, (c) drive frequencies, and (d) preload forces. Unless otherwise specified, the substrate width tested was 2 mm, the drive signal was a 300 V, 1 kHz bipolar square wave, and $F_{preload} <$ 0.25 N.}
	\label{fig:release_time_results}
\end{figure}

\subsection{Release Time Results and Discussion} \label{subsec:release_time_results}
Fig. \ref{fig:release_time_results} shows the release times of our EA-ICE clutches across a total of 311 test runs over different drive voltages and frequencies, substrate widths, and preload forces. The best release time was 874.6 \textmu s ($\sigma = $ 31.2 \textmu s) for a 2 mm wide substrate, a low preload force (averaged over test runs with $F_{preload} < 0.25$ N), and a 225 V, 1 kHz drive signal. Across all 311 runs in our dataset, the average release time was 1.13 ms ($\sigma =$ 0.36 ms), and 39.5\% of the runs had release times $< 1$ ms.

Our $1.11 - 3.33$ cm$^2$ EA-ICE clutches have much faster release times than other centimeter-scale EA clutches are reported to have in prior literature. Hinchet and Shea \cite{Hinchet_Shea_2020} reported release times around 15 ms for a 3 cm$^2$ parallel plate clutch, while Diller et al. \cite{Diller_Majidi_Collins_2016} measured 23.6 ms release times for a 80 cm$^2$ parallel plate clutch. The best EA-ICE clutch result we found was by Cao et al. \cite{Cao_Gao_Guo_Conn_2019}, who measured release times of 100 ms for a 24.6 cm$^2$ clutch adhering to a cardboard substrate. Our results fall behind the release times of sub-millimeter-scale EA clutches like MEMS gap closing actuators and relays, which can take advantage of their low mass and smaller relative damping forces to achieve microsecond-scale release times \cite{Rauf_Contreras_Shih_Schindler_Pister_2022, Nielson_Olsson_Resnick_Spahn_2007, Verger_Pothier_Guines_Blondy_Vendier_Courtade_2012}.

In Fig. \ref{fig:release_time_results}(a), linearly interpolating the experimental data gives a slope -0.72 \textmu s V$^{-1}$ compared to our model's predicted slope -0.25 \textmu s V$^{-1}$. 300 V drive voltages caused, on average, 20.7\% smaller release times than 125 V drive voltages did. In Fig. \ref{fig:release_time_results}(b), linearly interpolating the experimental data gives a similar slope 143.5 \textmu s mm$^{-1}$ compared to our model's predicted slope 142.3 \textmu s mm$^{-1}$. 2 mm wide substrates had 83.8\% smaller release times on average than 6 mm wide substrates did. In prior work, Diller et al. \cite{Diller_Collins_Majidi_2018} similarly observed that EA release times decrease with voltage and increase with substrate width.

In Fig. \ref{fig:release_time_results}(c), we observed a local minima for release times, with 2 kHz drive frequencies causing on average 44.2\% and 10.5\% lower release times than 1 Hz and 10 kHz drive frequencies respectively. We did not find a model that captures the low-frequency trend. Our original hypothesis was that our drive circuit might be leaking charge into the EA clutch, but Fig. \ref{fig:force_vs_voltage_vs_frequency} shows our clutch's shear force capacity did not increase at low frequencies. For drive frequencies above 2 kHz, however, our model did predict our experimental trend towards higher release times. At high frequencies, the EA force capacity decays, increasing the impact of slower mechanical dynamics in the release time. Notably, our experimental trends for release time line up very well with the trends in slip percentage from Fig. \ref{fig:force_vs_voltage_vs_frequency}(b), where we saw the percentage of force lost after a slip decrease significantly as the drive frequency increased from 10 Hz to 2 kHz but increase slightly as the drive frequency increased to 10 kHz. We hypothesize that because our EA clutches can release faster around 2 kHz, at those drive frequencies it can also more quickly dissipate the interfacial static charges that accumulate after a slip and recover faster.

In Fig. \ref{fig:release_time_results}(d), our model predicted that higher preload forces would cause lower release times, since as the preload force increases, the initial air gap is expected to decrease and the initial EA force is expected to increase. Because we initialized our load cell's position proportionally to $F_{shear,max}^{predicted}$, our model expected that the load cell would be initialized in a more deformed state, and thus that it would more quickly surpass static friction once the voltage is turned off. In practice, we measured larger release times at higher preload forces. We hypothesize the preload force caused our compliant dielectric film to wrap slightly around the substrate's edges, resulting in the EA pad also interfacing with the substrate's sidewalls.

Beyond the parameters varied in Fig. \ref{fig:release_time_results}, we believe our fast release times are helped by our use of a metallic substrate, which can quickly dissipate induced charge buildup, and our use of a fast push-pull drive circuit output stage, which can quickly sink charge away from the electrodes. Both of these have been used to good effect before for square EA clutches driven by low-frequency drive signals \cite{Hinchet_Shea_2022, Nakamura_Yamamoto_2017, Li_2021, Wei_Xiong_Dong_Wang_Liang_Tang_Xu_Wang_Wang_2023}, and we believe our exploration of different geometric form factors and high voltage drive signals enable the millisecond-scale release times seen in Fig. \ref{fig:release_time_results}. We also tested the extent to which our fitted multiplier $\lambda_{ea}^{fit}$ affected our release time simulation results in Supplementary Materials, Sec. S8, and we conclude it had a relatively minor effect for most testing conditions.

\begin{figure*}[!t]
	\centering
	\includegraphics[width=0.49\linewidth]{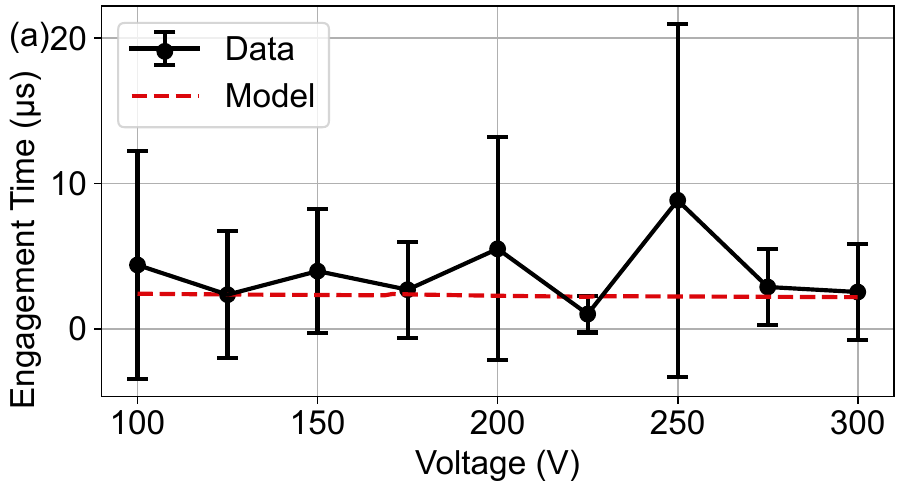}
	\includegraphics[width=0.49\linewidth]{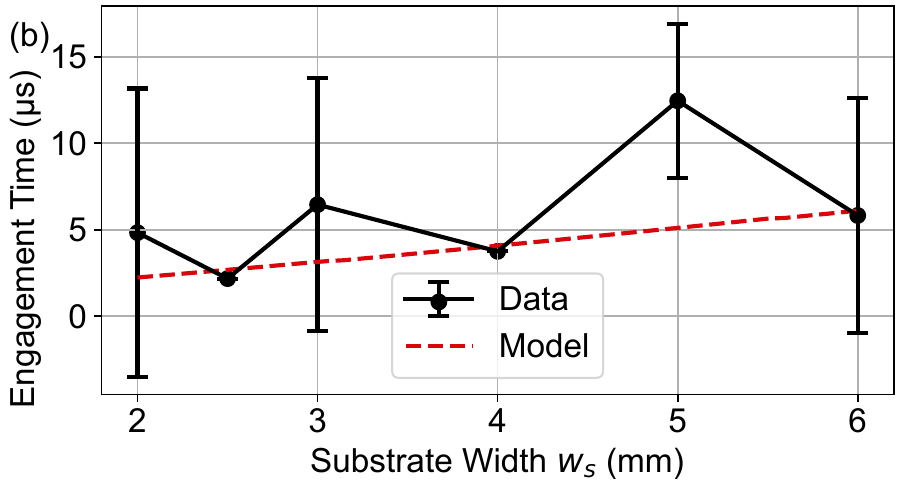}
	\includegraphics[width=0.49\linewidth]{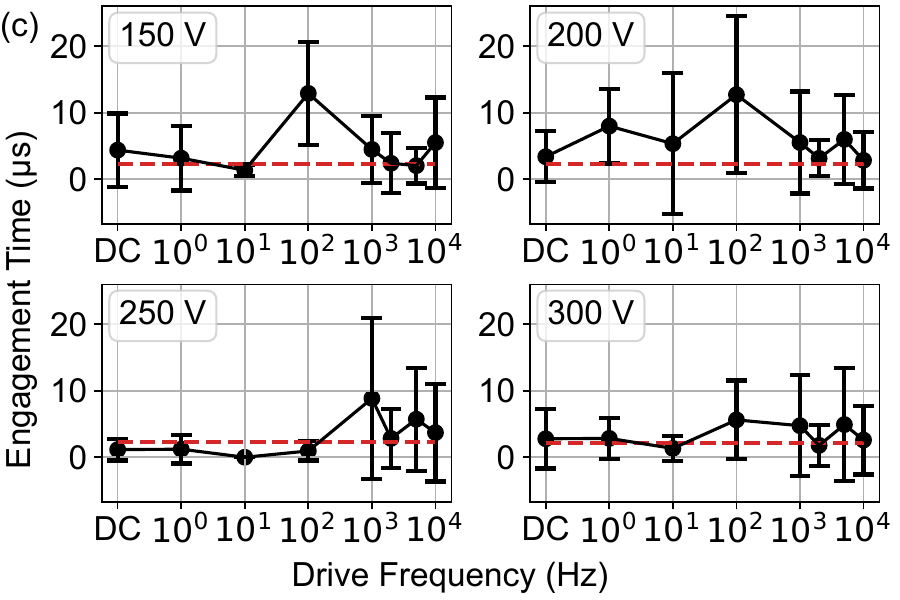}
	\includegraphics[width=0.49\linewidth]{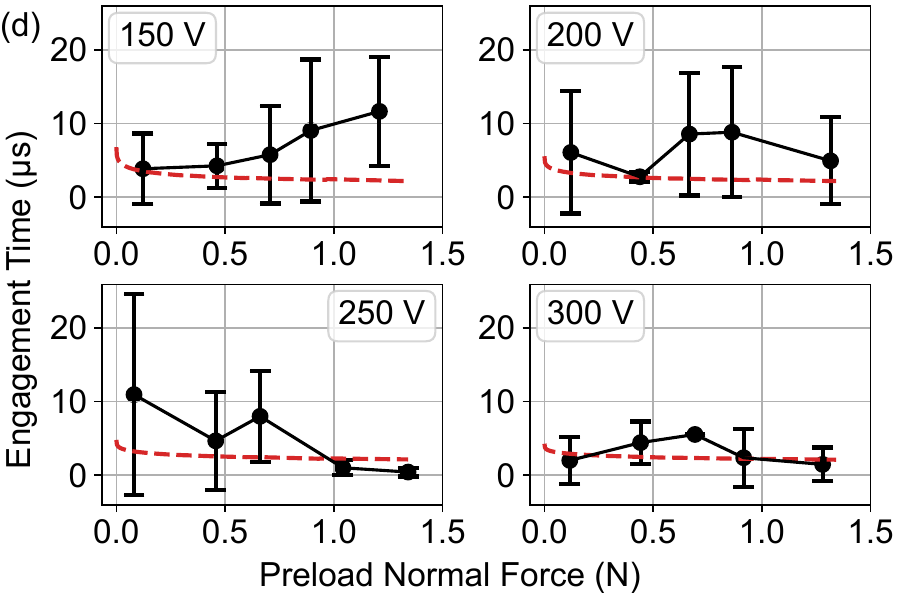}
	\caption{Engagement time averaged over at least 4 trials per condition for different (a) voltages, (b) substrate widths, (c) drive frequencies, and (d) preload forces. Unless otherwise specified, the substrate width tested was 2 mm, the drive signal was a 300 V, 1 kHz bipolar square wave, and $F_{preload} <$ 0.25 N.}
	\label{fig:engagement_time_results}
\end{figure*}

\subsection{Engagement Time Testing and Simulation Methods} \label{subsec:fts_engagement_time}
To measure the engagement time, we adapted a method from Diller et al. \cite{Diller_Collins_Majidi_2018}, shown in Fig. \ref{fig:force_test_setup}(d). First, we filtered load cell readings through a zero-phase 250 Hz low-pass Butterworth filter, which was chosen to be similar to prior work \cite{Krimsky_Collins_2024, Diller_Collins_Majidi_2018, Diller_Majidi_Collins_2016, Sirbu_Bolignari_DAvella_Damiani_Agostini_Tripicchio_Vertechy_Pancheri_Fontana_2022}. We then linearly interpolated the load cell readings after the voltage is enabled down to the average force over the 10 ms before the voltage is enabled. To minimize the impact of noise, we found linear interpolations for every data interval starting from 0.5 ms after the voltage is enabled and ending between 2-20 ms after the voltage is enabled. After filtering to only fits with $R^2 > 0.8$, we chose the smallest positive interpolated intersection to be the engagement time. For some tests, there would be no positive values due to noise; in these cases, we rounded the run's measured engagement time up to 0.

When using the model to predict engagement time, we solved the initial value problem defined by the coupled equations of motion Eqs. \ref{eq:eom_dielectric_z} and \ref{eq:eom_substrate_z} for $T_{air}$ as a function of time. For each experimental run, we extracted a single piece of information to simulate its dynamics, estimating $F_{preload}$ by measuring the kinetic friction on the substrate during the first 3 sec before the voltage is enabled and applying Eq. \ref{eq:F_preload}. We then solved Eq. \ref{eq:eom_dielectric_z} to predict the initial air gap before the voltage is enabled $T_{air}(t=0)$ and the final air gap after the voltage saturates $T_{air}(t \to \infty)$. The initial condition was set to $[T_{air}, \dot{T}_{air}] = [T_{air}(t=0), 0]$. The simulation was ended once the air gap traverses 0.5\% of the way from $T_{air}(t=0)$ to $T_{air}(t \to \infty)$. This threshold was set by our ADC's effective noise-free number of bits at our sampling rate and gain, and corresponds to when we predict the ADC can first notice the force increasing on the load cell after the voltage is enabled.

\subsection{Engagement Time Results and Discussion} \label{subsec:fts_engagement_time_results}
Fig. \ref{fig:engagement_time_results} shows the engagement times of our EA-ICE clutches across the same 311 test runs as in Sec. \ref{subsec:release_time_results}, showing in general that the EA clutch can engage within several tens of microseconds. This matches our hypothesis when modeling that EA actuators can be designed to engage about as quickly as it takes to polarize their dielectric (for comparison, our power supply had a 10\% - 90\% rise time of 8.3 \textmu s, and our dielectric had a polarization relaxation time of $\tau = 2.82$ \textmu s). Across all 311 test runs, the average engagement time was 4.78 \textmu s ($\sigma$ = 7.07 \textmu s), and 89.7\% (279/311) of our measured engagement times were $<$15 \textmu s. We observed significant run-to-run variance in our engagement times, which is not uncommon in the literature given the stochastic nature of the interaction between dielectric asperities and the substrate \cite{Diller_Collins_Majidi_2018, Zhang_Gonzalez_Guo_Follmer_2019}. For this reason, we focus less on the specific values of our measured engagement times and more on their general trends and order of magnitude.

Comparing results for our $1.11 - 3.33$ cm$^2$ EA-ICE clutches to prior literature, Hinchet and Shea \cite{Hinchet_Shea_2020} achieved engagement times of 5 ms for a 3 cm$^2$ parallel plate EA clutch, while Zhang et al. \cite{Zhang_Gonzalez_Guo_Follmer_2019} achieved an engagement time of 6.7 ms for an 0.96 cm$^2$ EA-ICE clutch adhering to a brass metal substrate. In 1957, Fitch \cite{Fitch_1957} achieved an engagement time of 150 \textmu s for a 38 cm$^2$ parallel plate rotary clutch using silicone oil as the dielectric and very precisely machined electrodes, which we believe helped reduce their interfacial surface roughness. Sub-millimeter scale EA devices like xerographic toner drums \cite{Hays_1991} and MEMS gap closing actuators \cite{Rauf_Contreras_Shih_Schindler_Pister_2022, Nielson_Olsson_Resnick_Spahn_2007} also often have microsecond-scale engagement times because, despite having masses an order of magnitude less than that of our system, their initial air gaps are often proportionally much larger than ours due to their fabrication tolerances.

While our engagement times are, on average, better than prior work on centimeter-scale EA clutches, we do not consider them particularly surprising. While the EA release time must account for mechanical damping and load cell dynamics, during engagement the measured shear friction should, theoretically, increase as soon as induced charges accumulate in the substrate, polarizing the dielectric and causing it to reduce the air gap and apply a normal force onto interfacial asperities. Because of our metallic substrate and fast push-pull drive circuit output stage, the electrical dynamics should occur on the order of microseconds, as described in Sec. \ref{subsec:electrostatic_dynamics}. Thus, unlike previous EA dynamics models that predict second-scale time constants due to Maxwell-Wagner interfacial polarization \cite{Chen_Liu_Wang_Zhu_Tao_Zhang_Jiang_2022, Nakamura_Yamamoto_2017, Chen_Zhang_Song_Fang_Sindersberger_Monkman_Guo_2020}, our model correctly matches the order of magnitude of our experimental results. While we are not able to directly apply our model to prior work, given our lack of contact models for their dielectric-substrate interfaces, we explore several hypotheses for why other centimeter-scale EA clutches might see much larger engagement times in Sec. \ref{sec:modeling_parameter_sweeps}.

Although our large run-to-run experimental variance makes direct numerical comparison difficult, our model's predicted engagement times also generally follow the trends of our experimental data. In Fig. \ref{fig:engagement_time_results}(a), linearly interpolating the experimental data gives a similar slope -0.0013 \textmu s V$^{-1}$ compared to our model's predicted slope -0.0012 \textmu s V$^{-1}$. Both Diller et al. \cite{Diller_Collins_Majidi_2018} and Zhang et al. \cite{Zhang_Gonzalez_Guo_Follmer_2019} similarly observed that higher voltages corresponded to lower engagement times. In Fig. \ref{fig:engagement_time_results}(b), linearly interpolating the experimental data gives a slope 1.15 \textmu s mm$^{-1}$ compared to our model's predicted slope 0.98 \textmu s mm$^{-1}$. Notably, although the EA normal force is linearly proportional to the substrate width, we see a positive correlation between engagement time and substrate width instead of a negative one because the damping force scales as the cube of substrate width. In Fig. \ref{fig:engagement_time_results}(c), as predicted by our model, we did not observe a consistent trend between engagement time and drive frequency because our measured engagement times were less than the period of any of our tested drive signals. In Fig. \ref{fig:engagement_time_results}(d), our model predicted a small increase in engagement time for zero preload normal force, but it quickly steadied for higher preload forces once the initial average air gap became small enough that electrical dynamics, rather than mechanical dynamics, dominated our simulation of engagement time. We observed no consistent experimental trend between our measured engagement times and applied preload forces.

\section{Simulating Parameter Sweeps for Engagement and Release Time} \label{sec:modeling_parameter_sweeps}
To inform future EA clutch designs, we extrapolated our dynamics model to plot its predicted engagement and release times across a wide range of design parameters. 

\begin{figure}[!t]
	\centering
	\includegraphics[width=0.49\columnwidth]{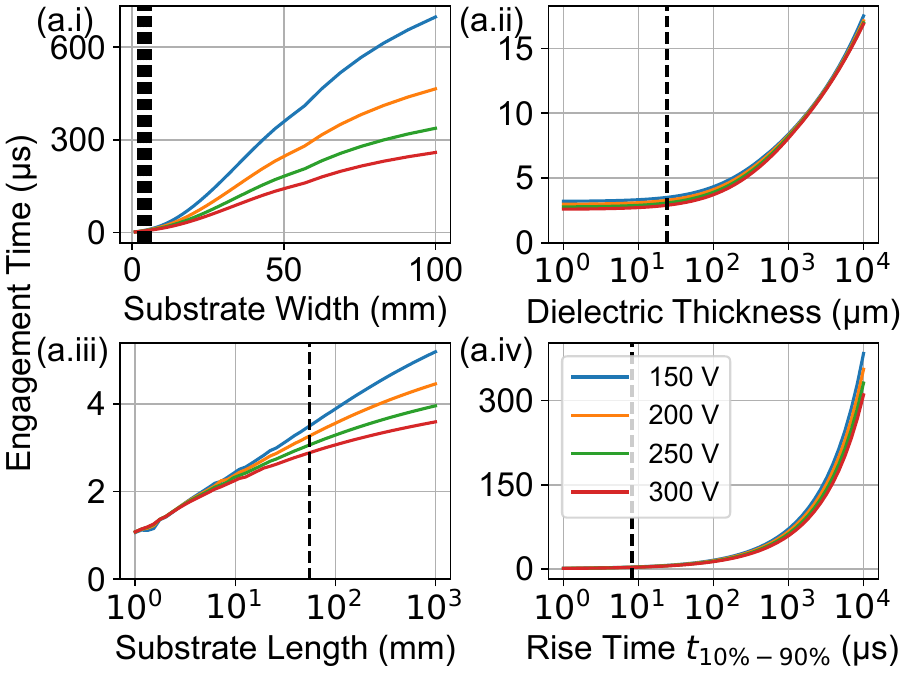}
	\includegraphics[width=0.49\columnwidth]{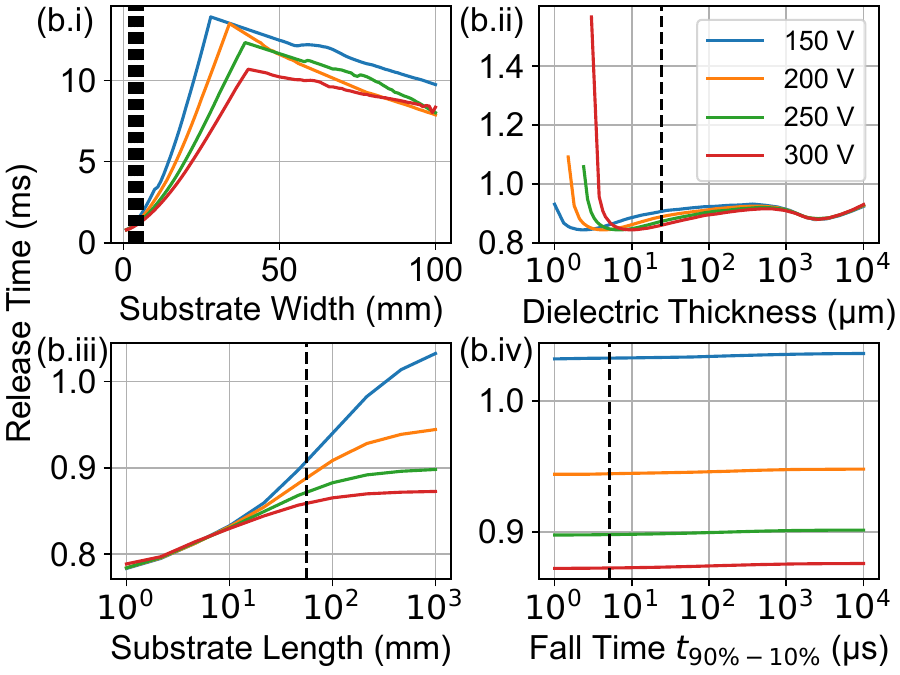}
	\caption{Our model's predicted (a) engagement times and (b) release times while varying voltage together with (i) substrate width, (ii) dielectric thickness, (iii) substrate length, and (iv) the 10\% - 90\% rise time or 90\% - 10\% fall time of our high voltage drive circuit. The black dotted lines indicate the values tested in this paper. Unless otherwise specified, the substrate width tested was 2 mm, the drive signal was a 300 V, 1 kHz bipolar square wave, $F_{preload} =$ 0.125 N, and $F_{shear}(t_r) / F_{shear,max}$ = 0.8.}
	\label{fig:sim_engage_time_vs_params}
\end{figure}

Fig. \ref{fig:sim_engage_time_vs_params}(a) shows our model's predicted engagement time over different drive voltages, substrate widths and lengths, dielectric thicknesses, and drive circuit rise times. As shown, the substrate width, drive circuit rise time, and voltage all have very large effects on engagement time, while the dielectric thickness and substrate length have much smaller effects. The substrate's width affects engagement dynamics much more than its length does because the damping force increases as the cube of width but linearly with length. We also swept a variety of other parameters that did not have a large impact on the predicted engagement time, largely because our actuators engage so quickly that electrical and mechanical dynamics all change on a similar time scale. Our model predicted that the dielectric constant $\kappa$ would have minimal effect on engagement time, because when the voltage first rises (i.e., when $T_{air}$ is the largest) the $\kappa^2$ terms in the numerator and denominator of Eq. \ref{eq:electroadhesive_normal_force} roughly cancel out. The preload force also had little effect in our predicted and experimental results, which is promising for future work because we found a small preload significantly reduces wrinkling in the dielectric film.

Fig. \ref{fig:sim_engage_time_vs_params}(b) shows our model's predicted release time over the same design parameters. The substrate width matters the most for release time, followed by voltage and dielectric thickness. For larger substrate widths, after the rate of increase in our contact force $F_k(T_{air})$ at small air gaps outpaces that of our EA force $F_{ea}(T_{air})$, larger substrate widths are predicted by our model to see lower release times. The dielectric thickness has an interesting local minima in its graph as the dielectric's increasing mass compensates for the loss of EA force capacity when determining the initial air gap before release. Diller et al. \cite{Diller_Collins_Majidi_2018} similarly observed higher release times at larger substrate widths and lengths, lower voltages, and in an intermediate dielectric thickness range for parallel plate EA clutches, but to our knowledge this is the first time these trends have been replicated by a physics-based model. The high voltage drive circuitry's fall time only has a small positive correlation with release time, as it slightly delays when the load cell can overcome static friction and because slower mechanical dynamics predominantly influence the release time. Although not shown in Fig. \ref{fig:sim_engage_time_vs_params}(b), because the electroadhesive force is proportional to the square of both voltage and the dielectric constant $\kappa$, the model similarly predicts that the dielectric constant has a negative correlation with release time. This prediction matches experimental results by Cao et al. \cite{Cao_Gao_Guo_Conn_2019}.

These simulations help establish important design tradeoffs for EA clutches. Because the electroadhesive force is proportional to substrate area, long, narrow EA clutches can achieve much faster engagement and release times while achieving the same force capacity. Diller et al. \cite{Diller_Collins_Majidi_2018} discovered a similar principle experimentally, when they found that cutting slits into a large EA pad significantly improved release time without harming shear force capacity, and we expect future timing-focused clutch designs will further explore the design space. Improved dynamics also result in faster recovery after slip, opening new uses for EA clutches in applications requiring robustness. The high voltage drive circuit's rise time also makes a large difference in an EA clutch's engagement time, which can motivate important decisions over the choice of high voltage amplifier and output stage needed for a project.

\section{Conclusions and Future Work} \label{sec:conclusion}

In this paper, we developed a dynamics model for EA-ICE clutches loaded in shear against metallic substrates. As a part of this model, we presented a simple calibration procedure to estimate the interfacial contact mechanics between the dielectric and substrate, bypassing the time and machine cost associated with traditional surface profilometry. We also integrated a Cole-Cole dielectric polarization model, proposing that dielectrics can be charged and discharged within microseconds rather than the seconds to minutes commonly assumed in prior modeling work. We then experimentally measured EA shear force capacity, release time, and engagement time over a variety of device parameters and compared our experimental trends to our model's predictions with good conformity. 

We conclude that the drive frequency, voltage, and substrate width to length aspect ratio significantly impact EA release times. While previous clutches in the literature have focused on square or circular patterns at low drive frequencies (DC or $\leq$10 Hz AC), our simulations and experimental results show that longer, narrower clutches at higher drive frequencies can provide a better trade-off between shear force capacity and release time. In our setup, 1-2 kHz drive frequencies achieved the highest shear force capacities, lowest release times, and best slip recoveries, but we expect this optimal range will vary for each EA clutch design. We also find, both experimentally and in simulation, that our use of a metal substrate and fast push-pull high voltage output stage significantly improved EA engagement time, allowing our EA clutches to engage about as fast as their dielectric can polarize across a wide range of geometric and operating parameters.

Our current work has some drawbacks, most notably in its simplified contact mechanics model and handling of parasitics. Future work should investigate more advanced contact models, integrating the substrate's surface energy and work function to understand what happens at the transition between frictional and peeling fracture mechanics, how the stochastic interactions between local asperities translates to the global shear force, and how surface roughness factors into the electroadhesive force as asperities compress into the soft dielectric. The ultimate ideal is a model with no fitted multiplier $\lambda_{ea}^{fit}(V)$. Our model also assumed that the substrate is a metal, in which charges can be induced much faster than in an insulating substrate. Future work should investigate faster and more precise tools to measure the engagement and release times across a variety of substrates. Optical interferometry \cite{Chen_Bergbreiter_2017} or digital image correlation \cite{Liu_Xie_Duan_Chen_Yuan_Ji_Huang_2025}, for example, could allow for higher sample rate, more spatially nuanced measurements of how asperities on the two surfaces adhere and release over time. In lieu of perfectly accurate physics-based models, an alternative approach could be a completely data-driven model for electroadhesion, similar to the one explored by Diller et al. \cite{Diller_Collins_Majidi_2018}. In addition, although we originally used lower preload forces to avoid plastically deforming the dielectric film, our finding that lower preload forces tend to improve release times opens interesting translational questions into how to best integrate EA clutches in the field. Prior studies have explored a variety of elastic preload mechanisms for different EA clutch geometries \cite{Diller_Collins_Majidi_2018, Vechev_Hinchet_Coros_Thomaszewski_Hilliges_2022, Hinchet_Shea_2022, Detailleur_Umans_Van_Even_Pennycott_Vallery_2021}, and additional topology optimization should explore optimal mechanisms that can minimize wear and release time while still ensuring good surface conformity.

We believe that improving the dynamics of electroadhesion opens new potential for high-bandwidth applications, such as soft robots and haptic user interfaces. EA clutches have already received attention among switchable adhesives for their low power, low cost, low mass, and scalable manufacturing processes. However, while micro-electromechanical systems have leveraged these advantages to create kHz-driven precision electrostatic actuators for centimeter-scale rockets and grippers \cite{Rauf_Contreras_Shih_Schindler_Pister_2022, Rauf_Kilberg_Schindler_Park_Pister_2020}, existing soft actuators generally operate under orders of magnitude less accuracy and speed \cite{Croll_Hosseini_Bartlett_2019}. Similarly, existing EA integrations into haptic user interfaces have historically faced limited bandwidth, with -3 dB frequencies on the order of tens to hundreds of Hertz \cite{Shultz_Peshkin_Colgate_2018, Leroy_Shea_2023}. By understanding more about how different geometric and operating parameters affect electroadhesion's response speed, we believe it can become a versatile, space- and power-efficient actuator for future robotic systems.

\section*{Acknowledgments}
This work is supported by the National Science Foundation award grant no. 2142782. The authors thank Wing-Sum Adrienne Law, Savannah Cofer, Danyang Fan, and Olivia Tomassetti for providing feedback on this paper. The authors also thank Hojung Choi and Kai Jun Chen for their insightful discussions on capacitive sensing and fracture mechanics, respectively.

\bibliographystyle{elsarticle-num} 
\bibliography{bibliography}

\end{document}


\begin{frontmatter}	
		\title{Supplementary Materials for Modeling the Dynamics of Sub-Millisecond Electroadhesive Engagement and Release Times}
		
		\author{Ahad M. Rauf}
		\ead{ahadrauf@stanford.edu}
		\author{Sean Follmer}
		\affiliation{organization={Department of Mechanical Engineering at Stanford University}, city={Stanford}, state={CA 94305}, country={USA}}
	\end{frontmatter}
	
	
	\tableofcontents
	
	\newpage
	
	\section{Literature Review Timing Data}
	The values plotted in Fig. 1(c) are provided with references in Table \ref{tab:fig1c_lit_review}.
	
	\begin{table}
		\caption{Data Plotted in Fig. 1(c) Literature Review (DNM = Did Not Measure)}
		\label{tab:fig1c_lit_review}
		\centering
		\begin{tabular}{|c||c|c|}
			\hline
			\textbf{Reference} & \textbf{Engagement Time (s)} & \textbf{Release Time (s)} \\
			\hline
			This Work & \textbf{$1.50\cdot 10^{-5}$} & \textbf{$8.75\cdot 10^{-4}$} \\
			\hline
			Hinchet 2020 \cite{Hinchet_Shea_2020} & $5.00\cdot 10^{-3}$ & $1.50\cdot 10^{-2}$ \\
			\hline
			Diller 2018 \cite{Diller_Collins_Majidi_2018} & $7.23\cdot 10^{-3}$ & $2.36\cdot 10^{-2}$ \\
			\hline
			Krimsky 2024 \cite{Krimsky_Collins_2024} & $2.40\cdot 10^{-2}$ & $3.00\cdot 10^{-2}$ \\
			\hline
			Li 2023 \cite{Li_Xiong_Ma_Yang_Ma_Tao_2023} & $1.55\cdot 10^{-2}$ & $4.35\cdot 10^{-2}$ \\
			\hline
			Wei 2023 \cite{Wei_Xiong_Dong_Wang_Liang_Tang_Xu_Wang_Wang_2023} & $5.00\cdot 10^{-2}$ & $3.70\cdot 10^{-2}$ \\
			\hline
			Yoder 2023 \cite{Yoder_Macari_Kleinwaks_Schmidt_Acome_Keplinger_2023} & $5.00\cdot 10^{-2}$ & $7.00\cdot 10^{-2}$ \\
			\hline
			Nakamura 2017 \cite{Nakamura_Yamamoto_2017} & $1.04\cdot 10^{-2}$ & $1.62\cdot 10^{-1}$ \\
			\hline
			Li 2021 \cite{Li_2021} & $2.50\cdot 10^{-2}$ & $3.24\cdot 10^{-1}$ \\
			\hline
			Thilakarathna 2024 \cite{Thilakarathna_Phlernjai_2024} & $2.70\cdot 10^{-1}$ & $2.70\cdot 10^{-1}$ \\
			\hline
			Cacucciolo 2022 \cite{Cacucciolo_Shea_Carbone_2022} & $3.00\cdot 10^{-1}$ & $3.00\cdot 10^{-1}$ \\
			\hline
			Mastrangelo 2023 \cite{Mastrangelo_Caruso_Carbone_Cacucciolo_2023} & $4.19\cdot 10^{-1}$ & $3.28\cdot 10^{-1}$ \\
			\hline
			Hinchet 2022 \cite{Hinchet_Shea_2022} & DNM & $6.50\cdot 10^{-2}$ \\
			\hline
			Cao 2019 \cite{Cao_Gao_Guo_Conn_2019} & DNM & $1.01\cdot 10^{-1}$ \\
			\hline
			Cacucciolo 2019 \cite{Cacucciolo_Shintake_Shea_2019} & DNM & $1.50\cdot 10^{-1}$ \\
			\hline
			Chen 1992 \cite{Chen_Sarhadi_1992} & $2.00\cdot 10^{-1}$ & $1.00\cdot 10^{1}$ \\
			\hline
			Jeon 1998 \cite{Jeon_Higuchi_1998} & $2.83\cdot 10^{-2}$ & $1.37\cdot 10^{1}$ \\
			\hline
			Chen 2022 \cite{Chen_Liu_Wang_Zhu_Tao_Zhang_Jiang_2022} & DNM & $5.00\cdot 10^{0}$ \\
			\hline
			Gao 2019 \cite{Gao_Cao_Guo_Conn_2019} & DNM & $3.60\cdot 10^{1}$ \\
			\hline
			Fitch 1957 \cite{Fitch_1957} & $1.50\cdot 10^{-4}$ & DNM \\
			\hline
			Zhang 2019 \cite{Zhang_Gonzalez_Guo_Follmer_2019} & $6.70\cdot 10^{-3}$ & DNM \\
			\hline
		\end{tabular}
	\end{table}
	
	\section{Modeling Parallel Plate Electroadhesive Clutches}
	While our derivations in the main paper focus on EA-ICE clutches, the final dynamics equations will almost identical for parallel plate EA clutches. The only difference will be in the formulation of our quasistatic electrostatic force model. In a parallel plate configuration with overlap area $A$, the capacitance between the two electrodes is given by:
	\begin{equation} \label{eq:capacitance}
		C = \frac{\kappa\varepsilon_0 A}{T_d + \kappa T_{air}}
	\end{equation}
	
	Rearranging this equation, we can estimate the air gap via:
	\begin{equation} \label{eq:t_air_from_capacitance}
		T_{air} = \frac{\varepsilon_0 A}{C} - \frac{T_d}{\kappa}
	\end{equation}
	
	The electric fields in the dielectric and air are $E_d = \frac{\sigma}{\kappa \varepsilon_0}$ and $E_{air} = \frac{\sigma}{\varepsilon_0}$, respectively. The potential difference $V$ between the two electrodes can be related to the electric fields and charge densities by:
	\begin{equation}
		V =  E_dT_d + E_{air}T_{air} = \sigma \left( \frac{T_d}{\kappa \varepsilon_0} + \frac{T_{air}}{\varepsilon_0} \right)
	\end{equation}
	
	The electroadhesive normal force is thus:
	\begin{equation} \label{eq:electroadhesive_normal_force}
		F_{ea} = \frac{\sigma A E_{air}}{2} = \frac{\kappa^2\varepsilon_0}{2} \frac{AV^2}{(T_d + \kappa T_{air})^2}
	\end{equation}
	
	Eq. \ref{eq:electroadhesive_normal_force} differs from a conventional parallel plate force derivation that ignores the effect of the air gap \cite{Rauf_Bernardo_Follmer_2023}:
	\begin{equation} \label{eq:force_electrostatic_noairgap}
		F_{es,T_{air} = 0} = \frac{\kappa\varepsilon_0}{2} \frac{AV^2}{T_d^2}
	\end{equation}
	
	Eq. 4 and Eq. \ref{eq:electroadhesive_normal_force} only differ by the $(V/2)^2$ factor. This is because the potential difference between electrodes in a parallel plate EA clutch is $V$, while the potential difference between electrodes and the substrate in an EA-ICE clutch is $V/2$. While Eq. \ref{eq:electroadhesive_normal_force} has been used for parallel plate form factors before for electrostatic wafer chucking \cite{Yoon_Choi_Hong_2023}, we include its derivation here to motivate future work to include the effect of the air gap in modeling quasi-static electroadhesive force density.
	
	\section{Cole-Cole vs. Maxwell-Wagner Polarization Model}
	\subsection{Adhering to Conductive Substrates}
	As discussed in Sec. 2.2.1, we use a Cole-Cole polarization model instead of a Maxwell-Wagner interfacial polarization model despite the latter's prevalence in prior electroadhesion modeling literature \cite{Nakamura_Yamamoto_2017, Chen_Liu_Wang_Zhu_Tao_Zhang_Jiang_2022, Chen_Zhang_Song_Fang_Sindersberger_Monkman_Guo_2020} (although, to our knowledge, no papers proposing this polarization model have yet numerically compared their model against experimental data). We explain this decision further here, inspired by discussion in Kremer and Schönhals \cite{Kremer_Schonhals_2003}. The time constant of Maxwell-Wagner interfacial polarization interfacing with a conductive substrate is:
	\begin{equation} \label{eq:time_constant_maxwell_wagner}
		\tau_{MW} = \varepsilon_0\left(\frac{\kappa T_{air} + T_d}{\sigma_dT_{air} + \sigma_0T_d}\right)
	\end{equation}
	where $\sigma_0$ and $\sigma_d$ are the electric conductivities of air and the dielectric, respectively. Assuming $\sigma_0 \approx 5 \cdot 10^{-16}$ S m$^{-1}$, $\sigma_d \approx 10^{-11}$ S m$^{-1}$ \cite{Zhu_Liu_Wang_2017}, $\kappa = 50$, $T_d$ = 24 \textmu m, and $T_{air} = 1$ \textmu m (just as an example), $\tau_{MW} = 65.5$ seconds. If $T_{air} = 10$ \textmu m instead, $\tau_{MW} = 46.4$ seconds. If, instead of P(VDF-TrFE-CFE), the dielectric is polyimide ($\kappa = 3.4$, $\sigma_d = 6.7 \cdot 10^{-16}$ S m$^{-1}$ \cite{kapton2025}), $\tau_{MW} = 22.0$ hours. These relaxation times are several orders of magnitude larger than most of the experimental results observed in Fig. 1(c) and Table \ref{tab:fig1c_lit_review}. Historically, Maxwell-Wagner polarization was first proposed in 1873 by Maxwell's ``A Treatise on Electricity and Magnetism'' \cite{Maxwell_1873}, referring to a dielectric particle suspended in an infinite dielectric medium, and as such it assumes the system is disconnected from a power supply that can rapidly source charge or sink charge away. We proceeded with the Cole-Cole model (for which, as fitted in Sec. 2.2.1, $\tau_0 = 2.82$ \textmu s) since most electroadhesive devices always stay connected to their power supply.
	
	As a technical note, we originally tried to fit the data in Fig. 3 to a Havriliak-Negami model, a more general form of the Cole-Cole equation which raises the denominator in Eq. 6 to a fitted exponent $\beta$. However, as expected by prior literature on relaxor ferroelectrics like P(VDF-TrFE-CFE) \cite{Solnyshkin_Kislova_2010}, the optimal fit found $\beta = 1$. This simplified the Havriliak-Negami model to Eq. 6 and signified that P(VDF-TrFE-CFE)'s dielectric dispersion is symmetric.	
	
	\subsection{Adhering to Insulating Substrates}
	\begin{figure}
		\includegraphics[width=\columnwidth]{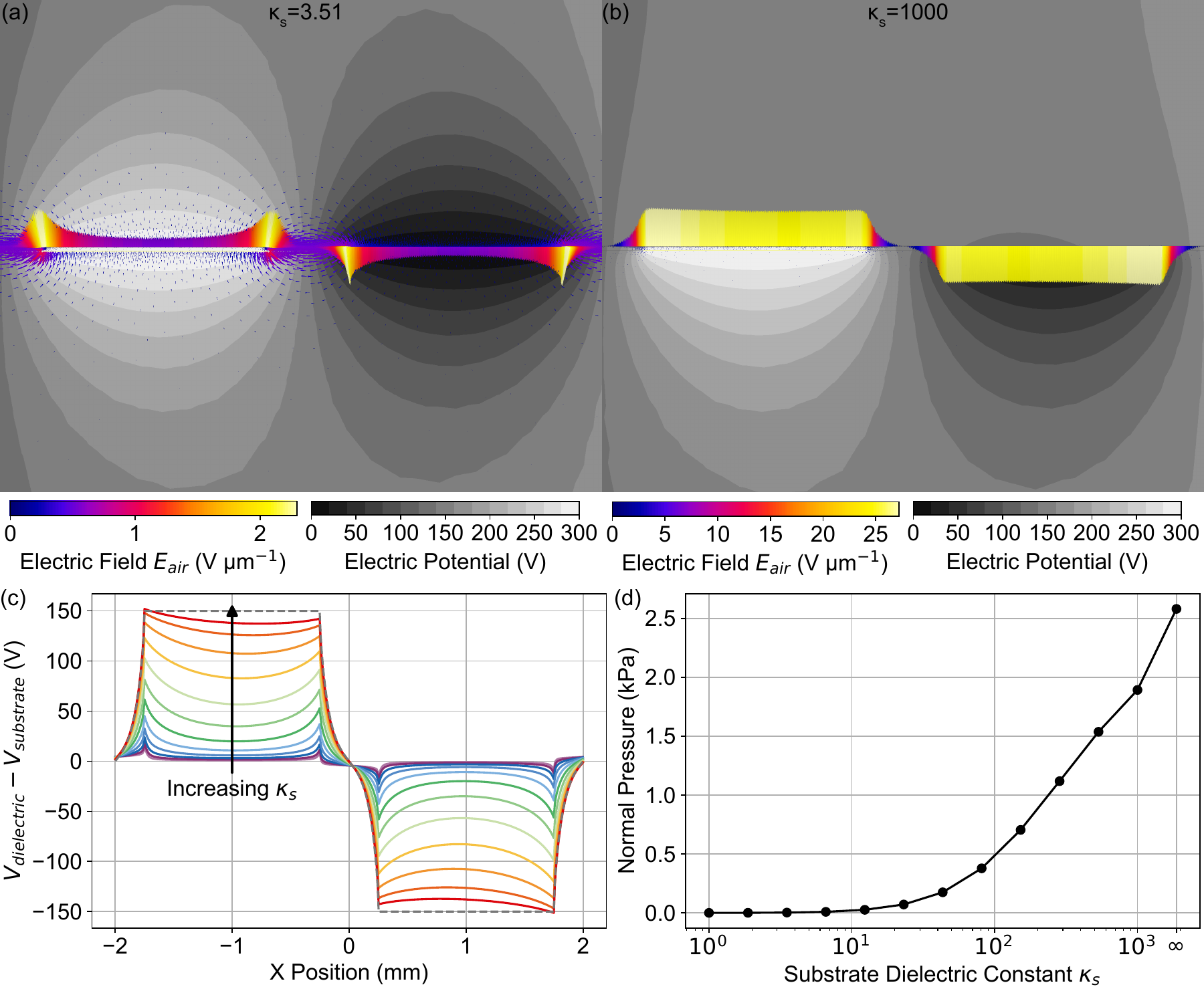}
		\caption{(a, b) Finite element analysis of the electric field (colored arrows) and potential (grayscale iso-contours) through an electroadhesive layer stack for a insulating substrate. The substrate's resistivity is 10$^{15} \Omega$m, and its dielectric constant $\kappa_s$ is (a) 3.51 and (b) 1000. The dielectric is a 24 \textmu m thick film dielectric with $\kappa = 50$, and the air gap is 5 \textmu m thick. Note that the electric field colormaps in (a) and (b) have different scales. (c) The potential difference between the dielectric's surface and the insulator's surface for different substrate dielectric constants, measured over the span of two adjacent electrodes. The grey dotted line shows the limit $\kappa_s \to \infty$ in metals. (d) The normal pressure on the substrate for different substrate dielectric constants.}
		\label{fig:fea_insulating_substrates}
	\end{figure}
	
	While this paper focuses on the dynamics of EA-ICE clutches adhering to conductive substrates, some insights also are applicable to future work modeling the dynamics of EA-ICE clutches adhering to insulating substrates. Fig. \ref{fig:fea_insulating_substrates} adapts the finite element simulation in Fig. 2(a) to use an insulating substrate with resistivity $\sigma_s = 10^{15}$ $\Omega$ m and dielectric constants $\kappa_s$ varying from 1 to 1000. Comparing the voltage iso-contours between Fig. \ref{fig:fea_insulating_substrates}(a) and Fig. \ref{fig:fea_insulating_substrates}(b), insulating substrates with low dielectric constants have substantially less uniform charge distributions and smaller gap voltages than insulating substrates with high dielectric constants (which roughly approximate the behavior of metals). Most insulating substrates in the literature, including plastics, wood, and paper, have dielectric constants between 3 and 4 \cite{Choi_Kim_Hwang_Kim_Park_Jang_Choi_Choi_Kim_Suhr_2020, Chen_Zhang_Song_Fang_Sindersberger_Monkman_Guo_2020, Chen_Liu_Wang_Zhu_Tao_Zhang_Jiang_2022, Berdozzi_Chen_Luzi_Fontana_Fassi_Molinari_Tosatti_Vertechy_2020}. However, at these dielectric constants, the average gap voltage is $<$2\% of the applied voltage (compared to 50\% for a conductive substrate), while the maximum potential difference inside the substrate itself is $>$97\% the applied voltage. This implies that the insulating substrate will retain charge in far more complex ways than simply how long it takes for the gap voltage to dissipate. While Maxwell-Wagner interfacial polarization primarily affects the gap voltage, we believe future work holds promise in investigating charge dissipation within the substrate itself (in addition to investigating the effect of other interfacial polarization mechanisms, such as electrode polarization \cite{AliAbbasi_Martinsen_Pettersen_Colgate_Basdogan_2024}). Chen et al. \cite{Chen_Huang_Tang_2017} and Koh et al. \cite{Koh_Kuppan_Chetty_Ponnambalam_2011} provide useful starter analytical derivations for the electric fields inside the insulator to help guide future work.
	
	\begin{figure}
		\includegraphics[width=\columnwidth]{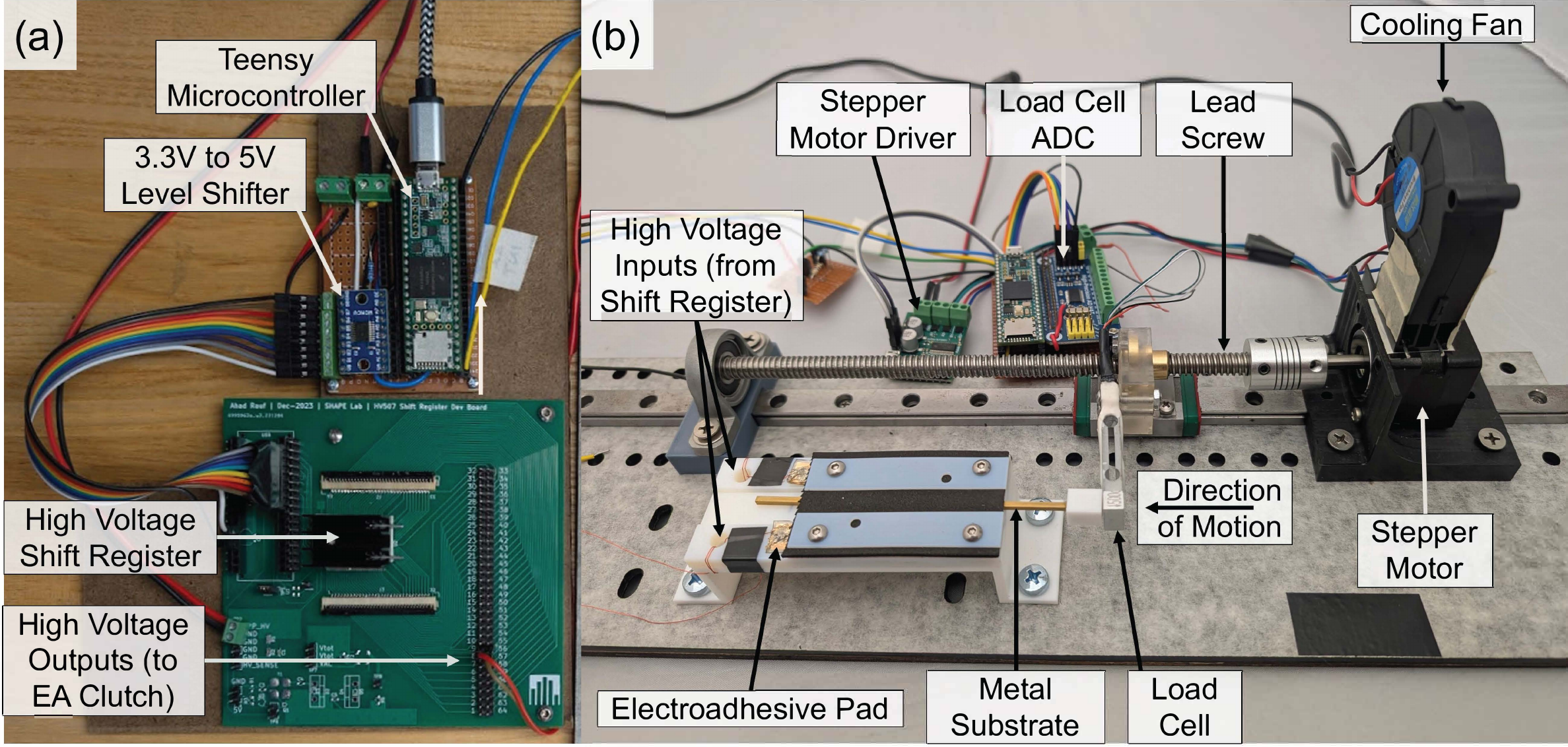}
		\caption{(a) The high voltage drive circuit board that controls the bipolar high voltage drive signals used in this study. (b) Experimental test setup for measuring EA shear force as the load cell on a motorized stage pushes against the substrate.}
		\label{fig:force_testing_setup_zoomin}
	\end{figure}
	
	\section{Experimental Test Setup Additional Details}
	This section provides additional implementation details for the experimental testing setup described in Sec. 3.1. Fig. \ref{fig:force_testing_setup_zoomin} shows additional photos of the high voltage drive circuit board and the tensile testing setup. We created a motorized linear slide by mounting a NEMA 14 hybrid bipolar stepper motor (SY35ST28-0504A, Pololu, USA) and a T8 lead screw to a linear rail. A 500g load cell (UXCell, China) was mounted to a linear slide block using screws and positioned so it can push the substrate. This setup, similar to other tensile testing machines used in \cite{Hinchet_Shea_2020, Diller_Collins_Majidi_2018, Berdozzi_Chen_Luzi_Fontana_Fassi_Molinari_Tosatti_Vertechy_2020, Kim_Lee_Bhuyan_Wei_Kim_Shimizu_Shintake_Park_2023}, uses the load cell's spring constant to decouple the motor and substrate's positions. The stepper motor was driven with a CNC motor driver (Tic 36v4, Pololu, USA) and was set to 1/256 microstepping and a current limit of 500 mA. To avoid timing variability over USB serial, data was first stored onto the microcontroller with timestamps before being uploaded to a computer for data processing. The Teensy 4.1 was modified with an additional 8 MB PSRAM to store all the data. Similar to \cite{Diller_Collins_Majidi_2018}, we chose not to use an Instron because of its embedded filtering and software delays, which prevent sub-millisecond measurement and time synchronization. In all tensile tests, we chose to push on the rigid substrate rather than pull our soft dielectric film, in line with the ISO 8295 standard for testing the static and kinetic friction coefficients of thin plastic films \cite{ISO8295} and the ASTM D3330 standard for 180$^\circ$ peel tests of thin adhesive lap joints \cite{ASTMD3330}.
	
	The P(VDF-TrFE-CFE) was etched using a quasi-continuous wave Diode Pumped Solid State (DPSS) UV laser cutter (Series 3500, DPSS Lasers Inc., USA). We used a small amount of isopropyl alcohol to suction the film to a polyimide-coated glass slide, which removed any wrinkles in the film before etching. The laser wavelength is 355 nm. The film was etched over 3 passes, with the laser set to a 120 kHz pulse frequency and moving at a speed 1200 mm/s. The film was then cut into our desired shape over 40 passes, with the laser set to a 40 kHz pulse frequency and moving at a speed 100 mm/s. The entire etch and cut process took about 550 seconds. After removing the EA pad from the UV laser cutter, enameled copper wire (PN155, Elektrisola, Germany) was adhered to each electrode using silver conductive epoxy (8331D, M.G. Chemicals, Canada) and allowed to cure for at least 6 hours at room temperature. The copper wire, silver epoxy, and gold electrodes collectively had a very low resistance ($< 1 \Omega$).
	
	The high voltage drive signal was powered by a DC-DC converter rated for up to 50 mA output current (AHV12V500V50MAW, Analog Technologies, USA) and a high voltage shift register with a half-bridge output stage (HV507, Microchip Technology Inc., USA). The rise time of our high voltage shift register is rated as 4 ns with a 20 pF capacitive load, which roughly corresponds to an output impedance of 200 $\Omega$. We found that the shift register could get very hot when powering high-frequency ($>$ 5 kHz) drive signals, so we adhered a heat sink and blower fan onto the shift register. To avoid timing uncertainty between the high voltage drive signals and stepper motor drive signals, the shift register was controlled using a second Teensy microcontroller, as shown in Fig. \ref{fig:force_testing_setup_zoomin}(a). This second microcontroller used interrupts to coordinate timing with the original Teensy, which handled the rest of the stepper motor timing and data logging.
	
	To measure the friction coefficient, we took the ratio between tangential force and normal force at three different normal force values across 10 trials, measuring $\mu_{static} = 0.188$ $(\sigma = 0.010$) and $\mu_{kinetic} = 0.154$ $(\sigma = 0.008)$ between the brass substrate and P(VDF-TrFE-CFE); and $\mu_{static} = 0.281$ $(\sigma = 0.034)$ and $\mu_{kinetic} = 0.173$ $(\sigma = 0.008)$ between the brass substrate and the 3D printed base plate. 
	
	\begin{figure}
		\includegraphics[width=\columnwidth]{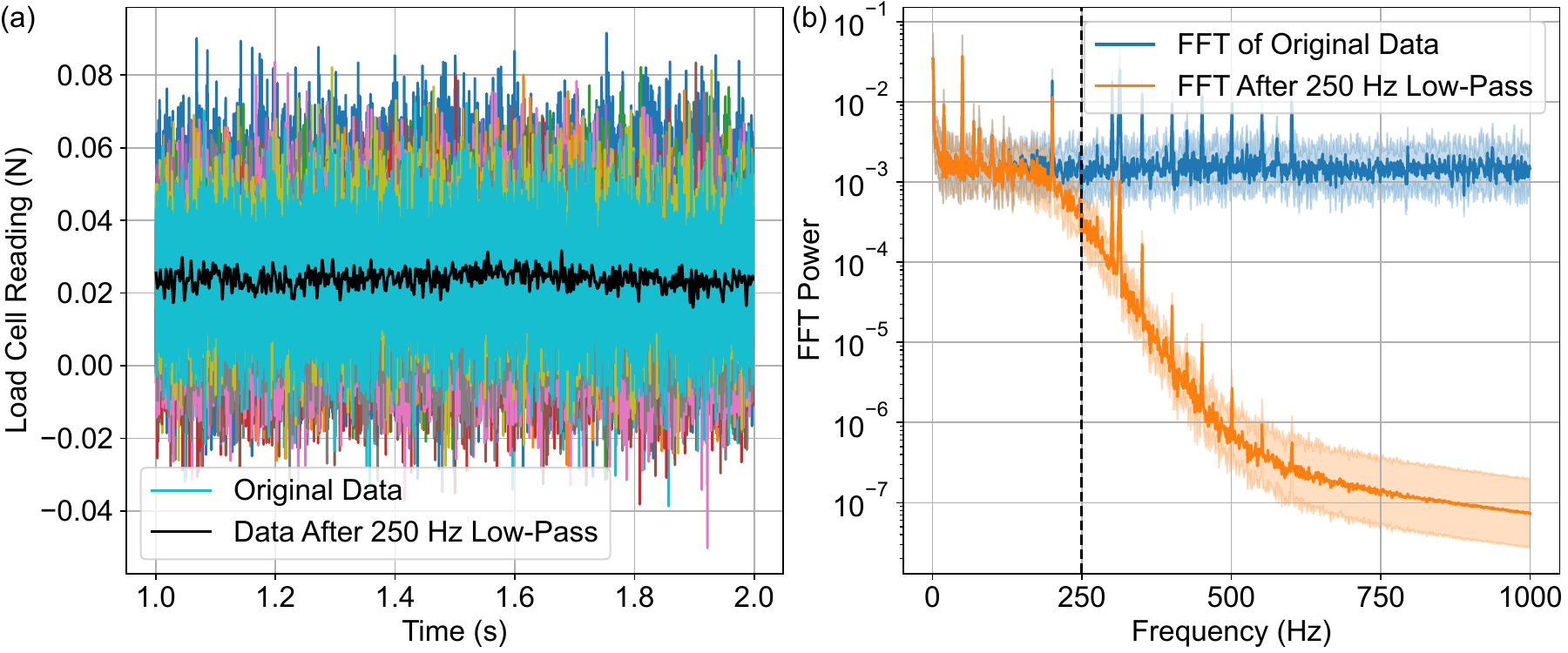}
		\caption{(a) Raw load cell readings for 20 experimental runs. Data was sampled between 1-2 sec into each experimental run, while the load cell is pushing the substrate but before the high voltage turns on. The black line shows the light blue raw data after a 250 Hz low-pass filter. (b) The mean and standard deviation of the FFT power spectrum before and after the 250 Hz low-pass filter (capped at 1 kHz for visualization purposes). As shown, the low-pass filter leaves lower-order signals intact but attenuates the higher-order vibration peaks between 300 - 600 Hz.}
		\label{fig:fts_noise_fft}
	\end{figure}
	
	Fig. \ref{fig:fts_noise_fft} shows the noise profile of our load cell reading while pushing the substrate with no high voltage applied, averaged over 20 experimental runs. The zero-phase 250 Hz low-pass Butterworth filter was chosen both to align with similar low-pass filters used in prior work \cite{Krimsky_Collins_2024, Diller_Collins_Majidi_2018, Diller_Majidi_Collins_2016, Sirbu_Bolignari_DAvella_Damiani_Agostini_Tripicchio_Vertechy_Pancheri_Fontana_2022} and to attenuate the higher-order vibration peaks in our data between 300 - 600 Hz.
	
	As a miscellaneous point, the high voltage bipolar square wave frequency could be quietly but audibly heard from the EA pad-substrate interface. For example, if the overall drive signal was 2 kHz, harmonics could be heard and visualized on a spectrum analyzer at 4, 8, 12, and 16 kHz. This could be useful in certain applications for remote monitoring or haptics.
	
	\begin{figure}
		\includegraphics[width=\columnwidth]{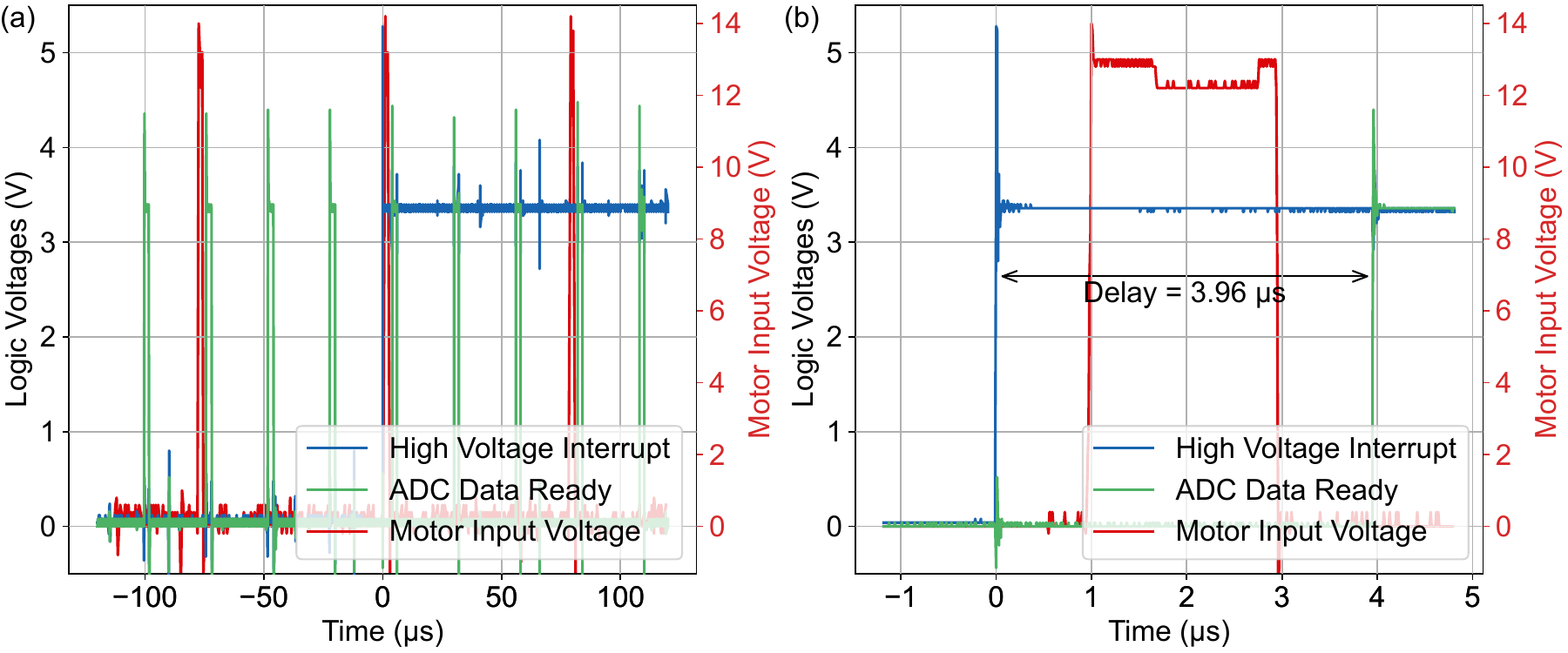}
		\caption{(a) Zoomed out view of the high voltage interrupt trigger (which instructs the HV 507 shift register to turn on the high voltage), the ADC's data ready interrupt, and the stepper motor's A1 input voltage during the 125 ms before and after the high voltage is turned on. The high voltage interrupt and ADC data ready curves are plotted using the left y-axis, and the motor input voltage curve is plotted using the right y-axis. (b) Zoomed in view of the plot in (a) during the microseconds before and after the high voltage is turned on.}
		\label{fig:adc_timing_diagram}
	\end{figure}
	
	\subsection{ADC Timing Diagram} \label{subsec:adc_sample_timing_diagram}
	For reliable microsecond-scale timing measurements, our microcontroller synchronized our ADC sample timings with the stepper motor step rate and the high voltage start time. Our ADC samples the load cell's Wheatstone bridge differential voltage at 921.6 kHz, and it saves this voltage using a sinc-weighted moving average filter. We recorded this differential voltage at a sample rate of 38.4 kHz. Meanwhile, the stepper motor in our tensile testing setup steps forward at 12.8 kHz based on our 0.5 mm s$^{-1}$ speed target, 1/256$\times$ microstepping ratio, 2 mm lead screw pitch, and 200 steps/revolution stepper motor. Fig. \ref{fig:adc_timing_diagram} shows the timing diagram of the high voltage interrupt trigger (which instructs the HV 507 shift register to turn on the high voltage), the ADC's data ready interrupt, and the stepper motor's A1 input voltage. For visual clarity, we only show one of the stepper motor's four drive signals (out of A1, A2, B1, and B2). The exact 3:1 ratio between the ADC sample rate and the stepper motor step rate was synchronized using our microcontroller. As shown in Fig. \ref{fig:adc_timing_diagram}(b), the stepper motor is set to start moving about 1 \textmu s after the high voltage turns on, and it settles within another 2 \textmu s. This settling time is also in line with the datasheet's specification. The ADC is set to have its first sample ready about 4 \textmu s after the high voltage turns on.
	
	\section{Tensile Test Sample Data}
\begin{figure}
		\includegraphics[width=\columnwidth]{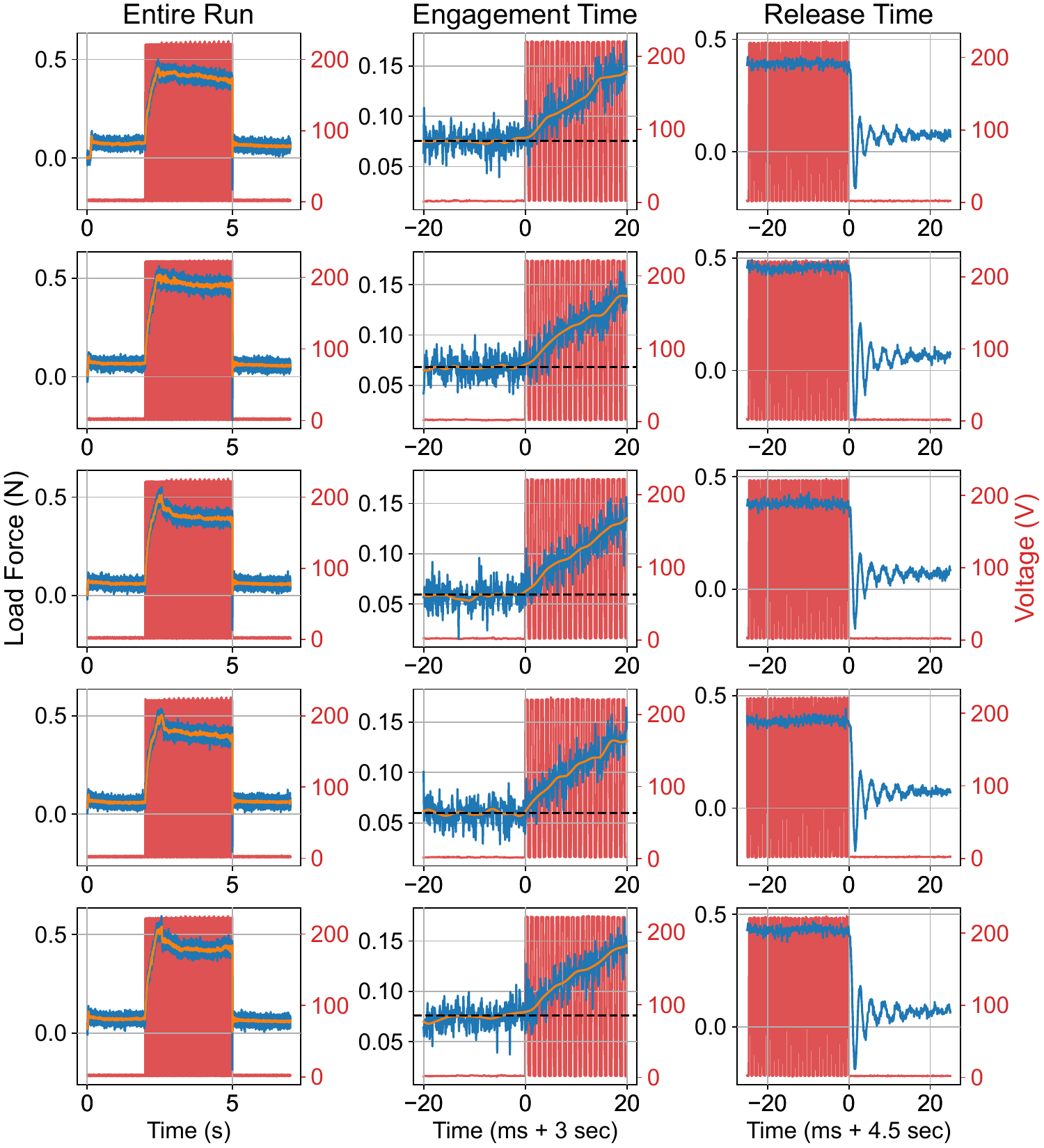}
		\caption{Five runs with a 2 mm wide substrate given a 225 V, 1 kHz drive signal and a low preload ($F_{preload} <$ 0.25 N). This condition had the smallest average release time among our test conditions. The first column shows the entire run, with raw load cell readings in blue, the low-pass filtered load cell readings in orange, and the voltage in red. The second column zooms into the engagement time. The third column zooms into the release time. The low-pass filtered readings were only used when measuring the engagement time in our results.}
		\label{fig:sample_data_225V_1kHz}
	\end{figure}

\begin{figure}
		\includegraphics[width=\columnwidth]{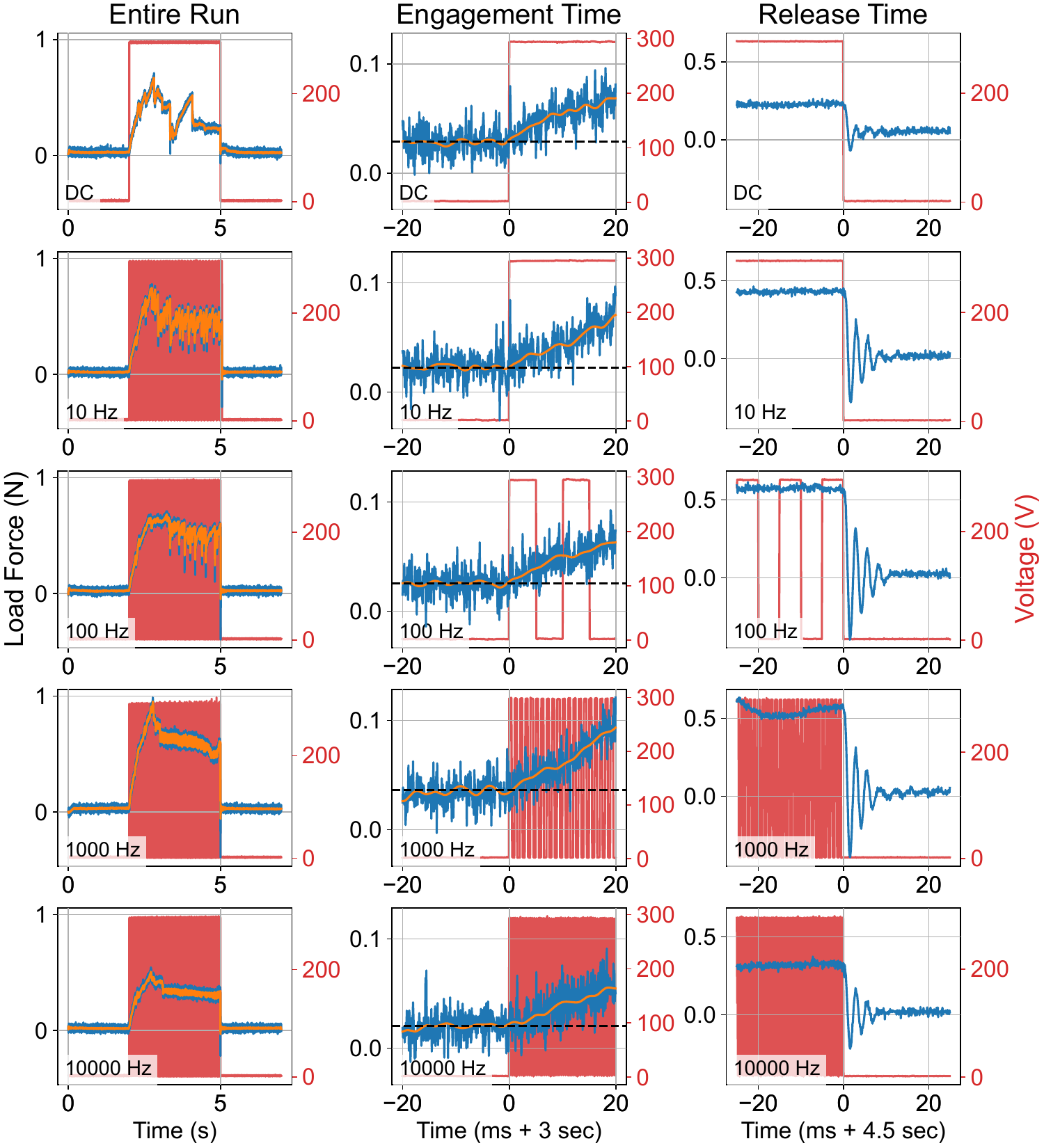}
		\caption{Tensile testing runs with a 2 mm wide substrate given a low preload ($F_{preload} <$ 0.25 N) and a 300 V drive signal at five different drive frequencies. The first column shows the entire run, with raw load cell readings in blue, the low-pass filtered load cell readings in orange, and the voltage in red. The second column zooms into the engagement time. The third column zooms into the release time. The low-pass filtered readings were only used when measuring the engagement time in our results.}
		\label{fig:sample_data_300V}
	\end{figure}

	We include comparisons of both the raw data and the data after the 250 Hz low-pass filter for 10 sample tensile testing runs below. Fig. \ref{fig:sample_data_225V_1kHz} shows five runs with a 2 mm wide substrate given a 225 V, 1 kHz drive signal and a low preload force ($F_{preload} <$ 0.25 N). This condition had the smallest average release time among our test conditions. Fig. \ref{fig:sample_data_300V} shows five runs with a 2 mm wide substrate given a 300 V drive signal and a low preload force ($F_{preload} <$ 0.25 N). Each row of Fig. \ref{fig:sample_data_300V} shows a different drive frequency, from DC to 10 kHz. As shown, although the 250 Hz low-pass filter shown in Fig. \ref{fig:fts_noise_fft} attenuates high-frequency noise, it preserves the data’s inherent trends.
	
	\section{Shear Force Capacity Measurement Additional Details}
	Experimentally, the relationship between shear force capacity and voltage varies widely between papers. While it is not uncommon for the relationship to be either linear ($F_{shear,max} \propto V$) \cite{Levine_Lee_Campbell_McBride_Kim_Turner_Hayward_Pikul_2023, Levine_Turner_Pikul_2021, Park_Drew_Follmer_Rivas-Davila_2020} or quadratic ($F_{shear,max} \propto V^2$) \cite{Chen_Bergbreiter_2017, Shultz_Peshkin_Colgate_2018, Yehya_Hussain_Wasim_Jahanzaib_Abdalla_2016}, the vast majority of papers we found lie on the spectrum $F_{shear,max} \propto V^n$. For example, Choi et al. \cite{Choi_Kim_Hwang_Kim_Park_Jang_Choi_Choi_Kim_Suhr_2020} had a best fit $F_{shear,max} \propto V^{0.686}$ between their EA-ICE pad and a paper substrate, and Diller et al. \cite{Diller_Collins_Majidi_2018} fitted $F_{shear,max} \propto V^{2.61}$ for their parallel plate EA clutch. The 1923 paper by Johnsen and Rahbek \cite{Johnsen_Rahbek_1923} that gave electroadhesion its name even found that $F_{shear,max} \propto V^5$ between a ``small cylinder of agate and a piece of thick tinfoil.''
	
	The literature has found a wide variety of quasistatic models for modeling the shear force capacity of electroadhesive actuators. Fracture mechanics-inspired models generally predict that the shear force capacity is linearly proportional to the voltage ($F_{shear,max} \propto V$) \cite{Levine_Lee_Campbell_McBride_Kim_Turner_Hayward_Pikul_2023, Levine_Turner_Pikul_2021}. Idealized frictional models that assume the air gap is either zero \cite{Rauf_Bernardo_Follmer_2023, Diller_Collins_Majidi_2018, Chen_Bergbreiter_2017, Schaller_Shea_2023} or uniform \cite{Shultz_Peshkin_Colgate_2018, Nakamura_Yamamoto_2017} generally predict that the shear force capacity is proportional to the voltage squared ($F_{shear,max} \propto V^2$). Despite the prevalence of other exponents in the literature, however, we have not seen an analytical model from first principles predicting experimental results on the spectrum $F_{shear,max} \propto V^n$.
	
	\begin{figure}
		\includegraphics[width=\columnwidth]{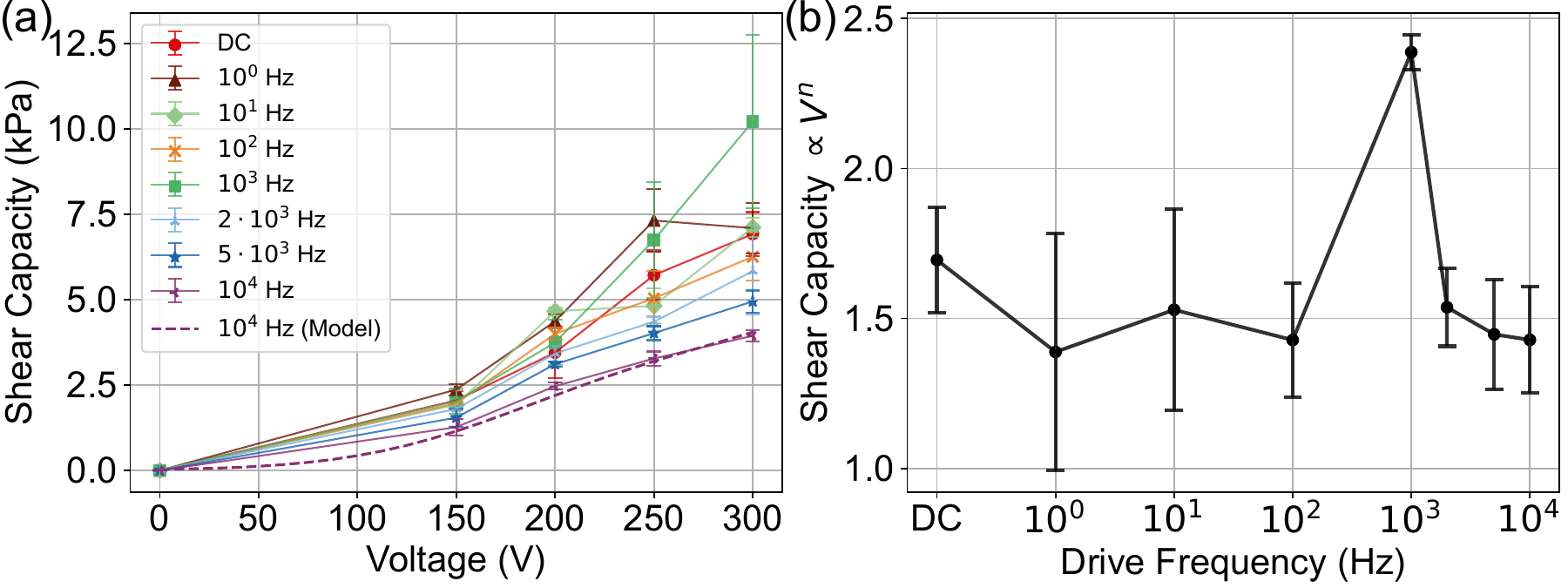}
		\caption{(a) Shear pressure capacity measured at different voltages and drive frequencies, averaged over at least 5 trials per condition. (b) The best fit exponent $F_{shear,max} \propto V^n$ for each of the lines in (a). One standard deviation of each fitted parameter is also shown.}
		\label{fig:force_vs_voltage_vs_frequency}
	\end{figure}
	
	Fig. \ref{fig:force_vs_voltage_vs_frequency}(a) shows the shear force capacity at different voltages and bipolar square wave AC drive frequencies across the same test runs as in Fig. 7(a), but plotting the shear force capacity against voltage instead. For each of the frequencies in Fig. \ref{fig:force_vs_voltage_vs_frequency}(a), we fit its curve against voltage to the equation $F_{shear,max}(V) = cV^n$. The best fit exponents $n$ (and one standard deviation) are shown for each frequency in Fig. \ref{fig:force_vs_voltage_vs_frequency}(b). The experimental data fit the curve $cV^n$ very well, with $R^2 = 0.979$ on average. The average fitted exponent across all frequencies was $n =$ 1.605 ($\sigma$ = 0.309).
	
	This non-integer exponent motivated us to explore a contact mechanics model for the current paper. The importance of surface roughness and contact mechanics for EA force capacity has been proposed before \cite{Johnsen_Rahbek_1923, Rajagopalan_Muthu_Liu_Luo_Wang_Wan_2022}, and the gap impedance has been characterized experimentally for different EA pads and substrates \cite{Ayyildiz_Scaraggi_Sirin_Basdogan_Persson_2018, AliAbbasi_Martinsen_Pettersen_Colgate_Basdogan_2024, Shultz_Peshkin_Colgate_2018}. However, we had not found an analytical model combining the electrical and mechanical equations into a full quasistatic electromechanical model in prior work, let alone using an electromechanical model to simulate dynamics. This is because, as said by Guo et al. \cite{Guo_Leng_Rossiter_2020}, ``it is non-trivial to derive analytical models that can accurately predict the theoretical electroadhesive forces due to the time-dependent nature of the EA effect and inhomogenous materials and electric fields existing in nature.'' Our model is not perfect, as shown by our use of a voltage-dependent fitted multiplier $\lambda_{ea}^{fit}(V)$ in Eq. 18, but we consider the design insights potentially interesting and in merit of future work.
	
	As an aside, we found that Eq. 4, which writes the electroadhesive force as a function of $T_{air}$, predicts a force far closer to our experimental results than the traditional parallel-plate capacitor equation (Eq. 5), which assumes $T_{air} = 0$. If we used Eq. 5 instead when predicting our EA clutch's shear force capacity, we found that on average $\lambda^{fit}_{ea} = 54.1$. This is much larger than the average $\lambda^{fit}_{ea} = 1.08$ needed when using Eq. 4, helping to validate our hypothesis that modeling the air gap is crucial to accurately characterizing electroadhesive force capacity.
	
	\section{Release Time Measurements Additional Details}
	We defined release time in the main paper as the 90\% fall time between when the voltage begins to fall and when the load cell falls 90\% of the way to its final steady-state value. While the 90\% fall time is more prevalent in the electroadhesion soft robotics literature \cite{Diller_Collins_Majidi_2018, Hinchet_Shea_2020}, the 10\% fall time is more prevalent when EA is used in micro-electromechanical systems, for example as a relay trigger mechanism \cite{Sakata_Komura_Seki_Kobayashi_Sano_Horiike_1999}. Papers integrating EA-ICE pads into grippers also often cite release time at the first sign of motion, rather than waiting for a full release \cite{Cacucciolo_Shea_Carbone_2022, Singh_Bingham_Penders_Manby_2016}. Here we compare results for the 90\% and 10\% fall times and discuss why we decided to report on the 90\% load cell fall times in the main body of the paper.
	
	\begin{figure*}[t]
		\centering
		\includegraphics[width=0.49\linewidth]{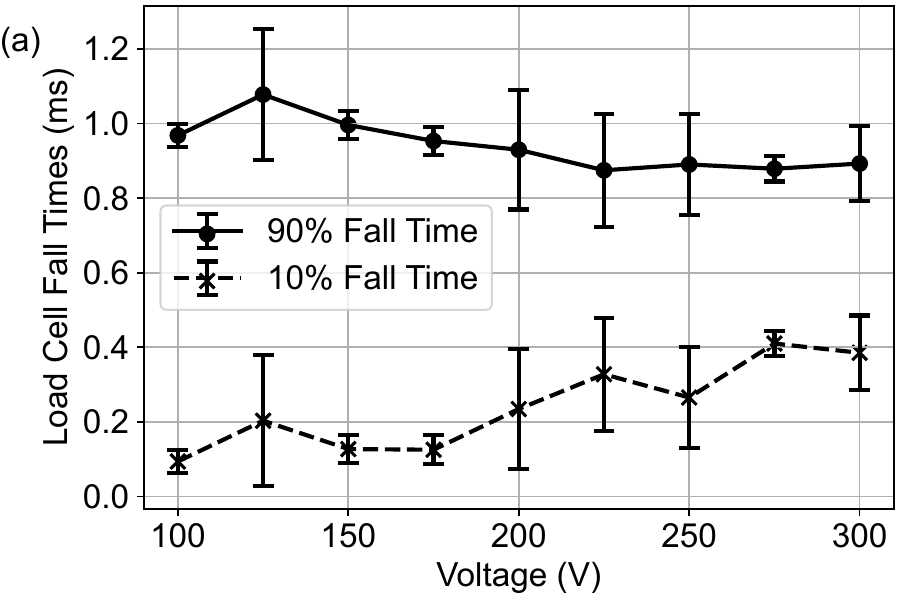}
		\includegraphics[width=0.49\linewidth]{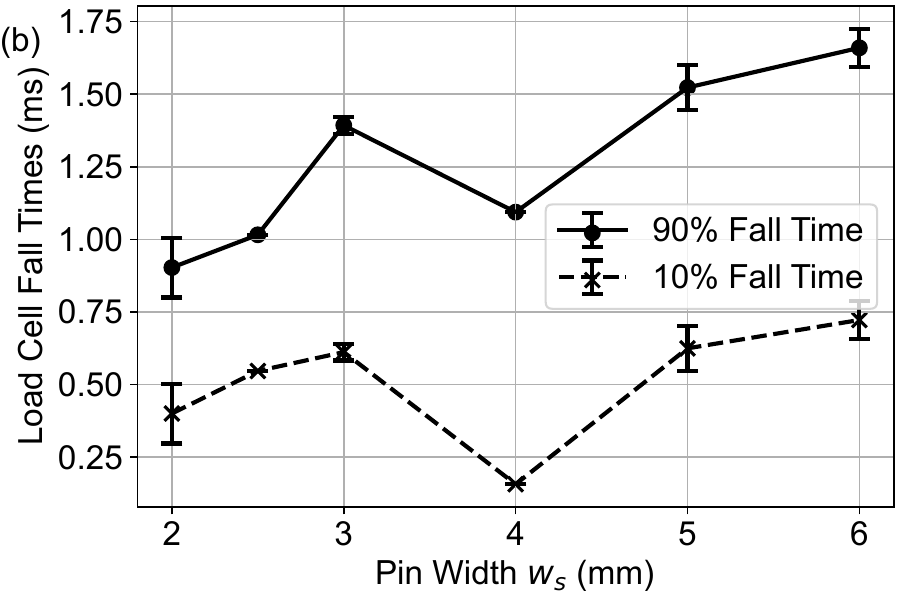}
		\includegraphics[width=0.49\linewidth]{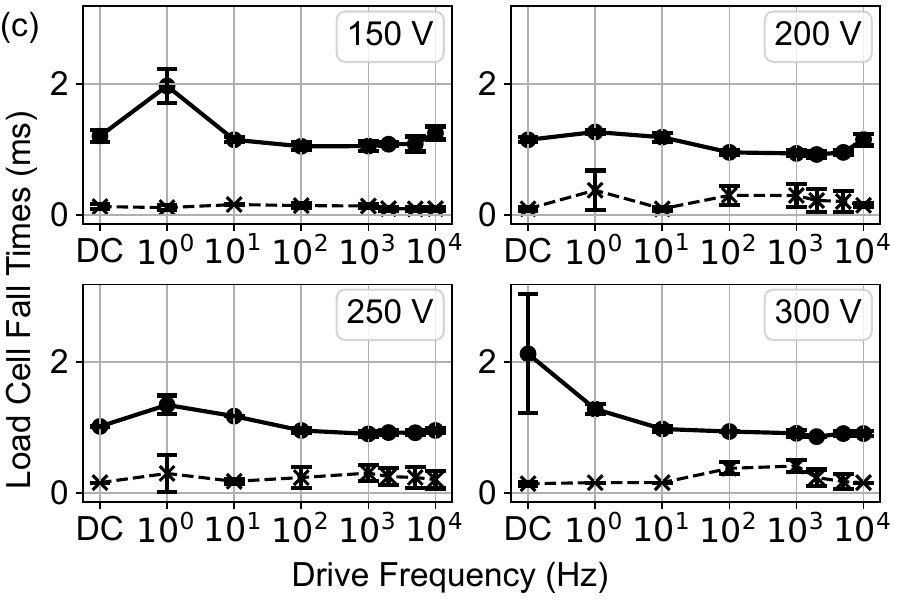}
		\includegraphics[width=0.49\linewidth]{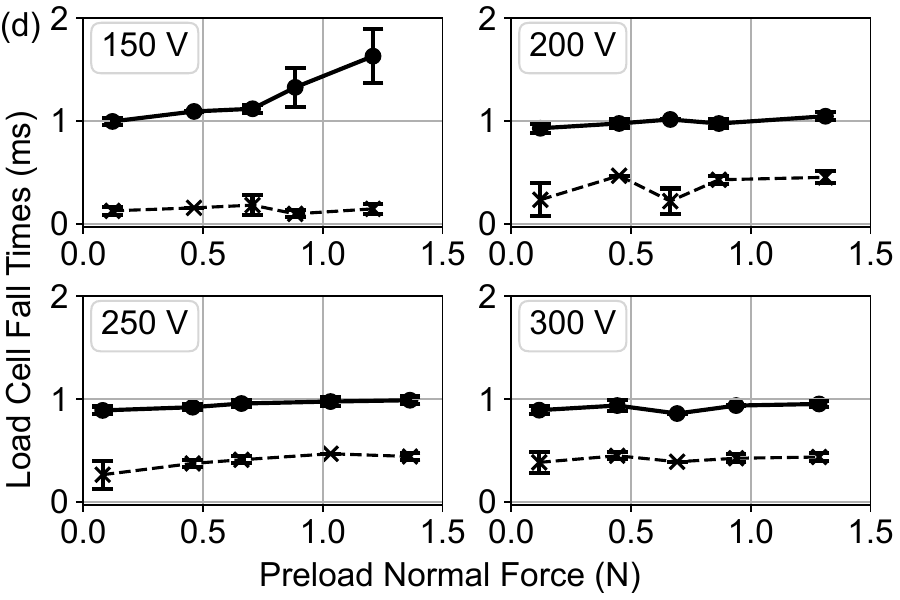}
		\caption{Comparison of the 10\% and 90\% load cell fall times, averaged over at least 4 trials per condition for different (a) voltages, (b) substrate widths, (c) drive frequencies, and (d) preload forces. Unless otherwise specified, the substrate width tested was 2 mm, the drive signal was a 300 V, 1 kHz bipolar square wave, and $F_{preload} <$ 0.25 N.}
		\label{fig:release_time_comparison_results}
	\end{figure*}
	
	Fig. \ref{fig:release_time_comparison_results} compares our 90\% and 10\% load cell fall time results. Across all of our 311 experimental runs, the average 10\% fall time was 286.0 \textmu s ($\sigma =$ 197.9 \textmu s), and the median 10\% fall time was 156 \textmu s. The average delay between the 90\% and 10\% fall times across all runs was 845.7 \textmu s ($\sigma =$ 399.3 \textmu s). This is larger than the load cell's quarter-period $\omega_{lc}/4 = 388.6$ \textmu s, implying that the residual EA forces might still be affecting release dynamics as the load cell settles. We assume that once the load cell settles, however, the residual EA forces have almost completely dissipated, since we noticed no significant trends in either the shear force capacity or timing results even when we conducted multiple experimental runs in immediate succession. This motivated our choice of a 90\% fall time metric to conservatively include the entire EA release process.
	
	Going into more detail, in Fig. \ref{fig:release_time_comparison_results}(a), we saw the 10\% load cell fall time increase with voltage, which we predict is because higher voltages cause the dielectric to compress more into the substrate and increase the electroadhesive force before release. The smaller initial air gap will take longer to return to equilibrium once the voltage is removed, resulting in larger 10\% fall times. When the load cell dynamics are accounted for during the 90\% fall time, however, this extra force also causes the load cell to bend more at higher voltages before release. The extra spring force thus causes the load cell to return to its origin position more quickly once it begins moving. The average delay between the 90\% and 10\% fall times decreased with voltage from 874.7 \textmu s at 100 V to 507.6 \textmu s at 300 V. Given the latter's proximity to the load cell's quarter-period, this implies that at low voltages, the load cell dynamics dominate our 90\% fall time metric, while at high voltages (which comprise the bulk of our experimental results) electroadhesive forces are the predominant factor in the 90\% fall time. In Fig. \ref{fig:release_time_comparison_results}(b), the 10\% fall times followed the same trend as the 90\% fall times. In Fig. \ref{fig:release_time_comparison_results}(c), we did not find any consistent trend among the 10\% release times with respect to frequency. We believe this means the interplay between EA release dynamics and load cell dynamics merit further work. In Fig. \ref{fig:release_time_comparison_results}(d), the 10\% fall times followed the same trend as the 90\% fall times. 
	
	\section{Release Time Simulations With and Without Fitted Multiplier $\lambda_{ea}^{fit}(V)$}
	
\begin{figure}[!t]
		\centering
		\includegraphics[width=0.49\linewidth]{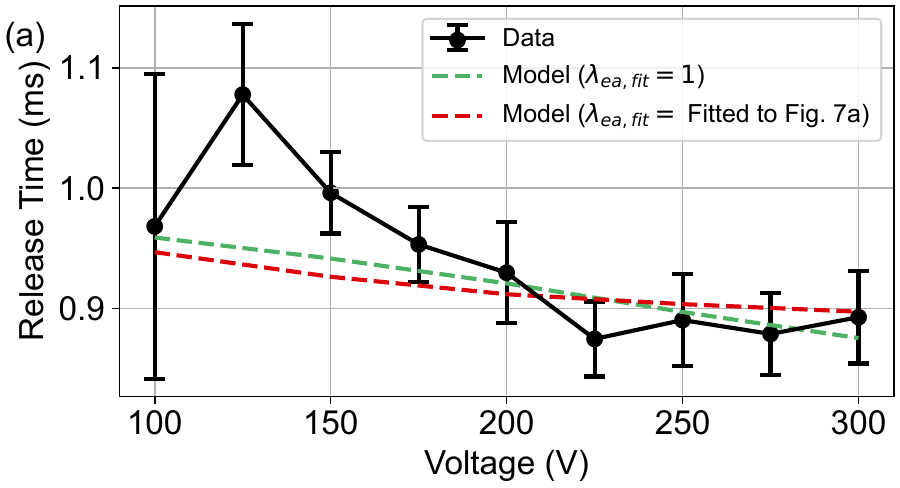}
		\includegraphics[width=0.49\linewidth]{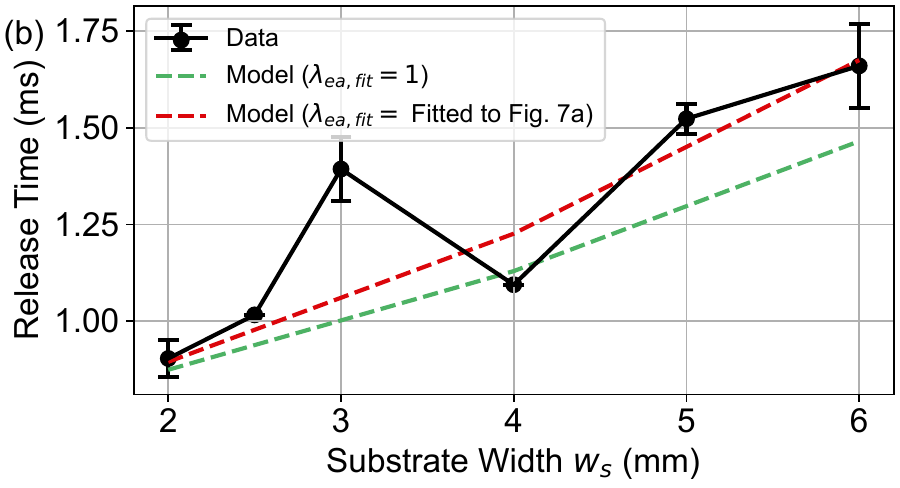}
		\includegraphics[width=0.49\linewidth]{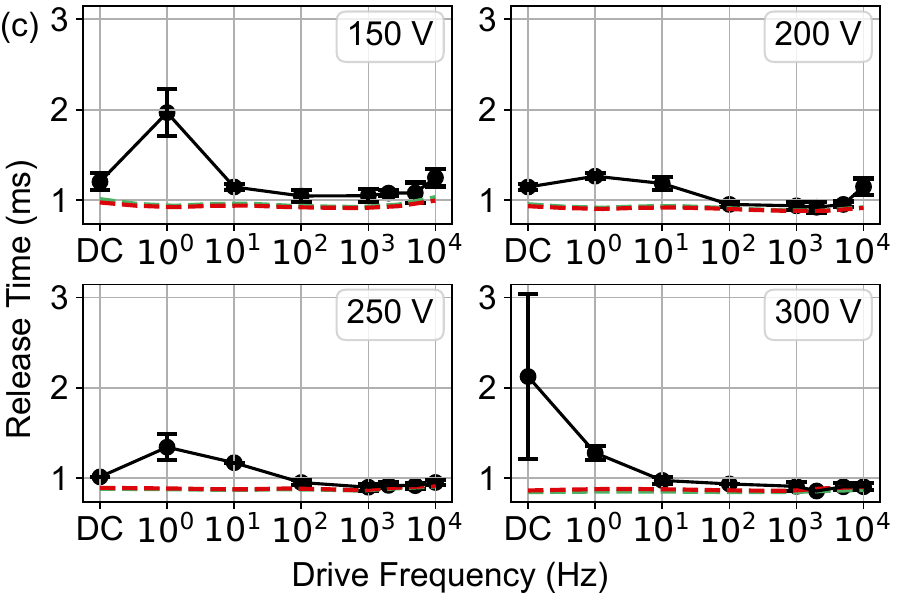}
		\includegraphics[width=0.49\linewidth]{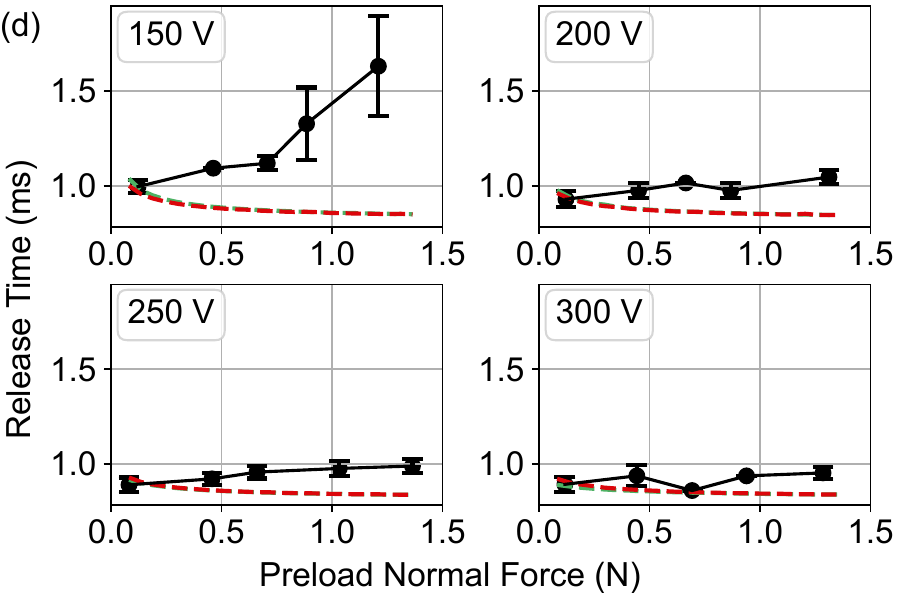}
		\caption{Release time averaged over at least 4 trials per condition for different (a) voltages, (b) substrate widths, (c) drive frequencies, and (d) preload forces. Unless otherwise specified, the substrate width tested was 2 mm, the drive signal was a 300 V, 1 kHz bipolar square wave, and $F_{preload} <$ 0.25 N. The red dotted line shows our model's predicted release time with our fitted multiplier $\lambda_{ea}^{fit}(V)$, and the green dotted line shows our model's predicted release time with an ideal $\lambda_{ea}^{fit} = 1$. In (c) and (d), the red and green model lines are almost identical.}
		\label{fig:release_time_results_Fesideal}
	\end{figure}

	To test the significance of our fitted multiplier $\lambda_{ea}^{fit}(V)$ on our dynamics simulations, we compared our model's predicted release time with and without the fitted multiplier in Fig. \ref{fig:release_time_results_Fesideal}. As shown, the simulated trends are similar with and without the fitted multiplier, although as expected using the fitted multiplier allows our predicted electroadhesive force, and thereby our dynamics simulation results, to more closely resemble the experimental results. Larger substrate widths were the only testing condition that showed significant difference between the two models, which could help inform future work investigating what physical phenomena result in our model's deviation from experimental results.
			
	\bibliographystyle{elsarticle-num} 
	\bibliography{bibliography_supplementary}